\documentclass[journal]{IEEEtran}

\ifCLASSINFOpdf
\else
\fi

\usepackage{amsmath,amssymb,amsfonts}
\usepackage{bm}
\usepackage{bbm}

\usepackage{enumitem}
\usepackage{float}
\usepackage{url}
\usepackage{multirow}
\usepackage{cite}
\usepackage{graphicx}
\usepackage{textcomp}
\usepackage[table]{xcolor}
\usepackage{booktabs}
\usepackage{adjustbox}
\usepackage{hyperref}
\usepackage{comment}
\usepackage{etoolbox}

\usepackage[caption=false,font=footnotesize]{subfig}

\usepackage{algorithm}
\usepackage{algpseudocode}

\makeatletter
\newcommand{\StatePar}[1]{%
  \State\parbox[t]{\dimexpr\linewidth-\ALG@thistlm}{\strut #1\strut}%
}
\renewcommand{\ALG@beginalgorithmic}{\footnotesize}
\makeatother

\DeclareMathOperator*{\argmax}{arg\,max}

\newcommand{\E}{\mathbb{E}}

\algrenewcommand\algorithmicrequire{\textbf{Input:}}
\algrenewcommand\algorithmicensure{\textbf{Output:}}
\algrenewcommand\algorithmiccomment[1]{\hfill\(\triangleright\) #1}

\definecolor{ARMORColor}{HTML}{FFF2CC}
\definecolor{ARMORPPColor}{HTML}{FCE4D6}

\setlist[itemize]{leftmargin=*,topsep=2pt,itemsep=1pt,parsep=0pt,partopsep=0pt}
\setlist[enumerate]{leftmargin=*,topsep=2pt,itemsep=1pt,parsep=0pt,partopsep=0pt}
\AtBeginEnvironment{thebibliography}{%
  \footnotesize
  \setlength{\itemsep}{0pt plus 0.2pt}%
  \setlength{\parsep}{0pt}%
  \setlength{\parskip}{0pt}%
}

\begin{document}

\title{ARMOR++: Agentic Reasoning for Method Orchestration and Reparameterization in Transferable Black-Box Attacks on Deepfake Detectors}

\author{
Christos~Korgialas,~\IEEEmembership{Graduate Student Member,~IEEE,}
Gabriel~Jun~Rong~Lee,
Dion~Jia~Xu~Ho,
Pai~Chet~Ng,~\IEEEmembership{Member,~IEEE,}
Xiaoxiao~Miao,~\IEEEmembership{Member,~IEEE,}
and~Konstantinos~N.~Plataniotis,~\IEEEmembership{Fellow,~IEEE}%
\thanks{Christos Korgialas is with the Department of Informatics, Aristotle University of Thessaloniki, Thessaloniki, Greece (e-mail: ckorgial@csd.auth.gr).}%
\thanks{Gabriel Jun Rong Lee and Pai Chet Ng are with the Infocomm Technology Cluster, Singapore Institute of Technology, Singapore (e-mail: 2301906@sit.singaporetech.edu.sg, paichet.ng@singaporetech.edu.sg).}%
\thanks{Dion Jia Xu Ho is with the Department of Applied Physics and Applied Mathematics, Columbia University, New York, NY, USA (e-mail: dh3065@columbia.edu).}%
\thanks{Xiaoxiao Miao is with the Division of Natural and Applied Sciences, Duke Kunshan University, Kunshan, China (e-mail: xiaoxiao.miao@dukekunshan.edu.cn).}%
\thanks{Konstantinos N. Plataniotis is with the Department of Electrical and Computer Engineering, University of Toronto, Toronto, ON, Canada (e-mail: kostas@ece.utoronto.ca).}%
}

\markboth{IEEE Transactions on Reliability, Vol.~XX, No.~XX, Month~2026}%
{Korgialas \MakeLowercase{\textit{et al.}}: ARMOR++: Agentic Orchestration of a Multi-Domain Primitive Set for Transferable Attacks on Deepfake Detectors}

\maketitle

\begin{abstract}
The reliability of deepfake detectors frequently degrades under black-box adversarial transfer, as these models often rely on fragile, architecture-dependent forensic cues. Existing transfer attacks often lack semantic awareness and struggle to maintain effectiveness under strict no-query constraints, particularly when perturbations are transferred from convolutional surrogates to transformer-based targets. To address these limitations, this paper introduces ARMOR++, a robust multi-agent framework designed for high-transferability deepfake evasion. The framework leverages the Qwen2.5-VL Vision-Language Model (VLM) to supply spatial semantic priors, while the Qwen3 Large Language Model (LLM) orchestrates primitive selection, adaptive hyperparameter reparameterization, and entropy-regularized perturbation mixing. By integrating five complementary primitives, spanning dense optimization, saliency-based methods, spatial transformations, frequency-domain perturbations, and block-structured modifications, ARMOR++ effectively targets heterogeneous inductive biases. Rigorous evaluation on the AADD-2025 benchmark demonstrates that ARMOR++ significantly outperforms existing agentic and non-agentic baselines across both low- and high-quality image regimes. Statistical analysis confirms a substantial gain in blind-target Attack Success Rate (ASR) over the state-of-the-art agentic baseline, with further performance advantages evidenced against non-agentic benchmarks and under robust defensive configurations. These findings highlight a significant residual reliability gap in current deepfake detector deployments and demonstrate the efficacy of agentic orchestration in identifying latent vulnerabilities.
\end{abstract}


\IEEEpeerreviewmaketitle

\section{Introduction}\label{sec:intro}

\IEEEPARstart{T}{he} rapid proliferation of generative artificial intelligence has transformed the production, dissemination, and forensic verification of visual media. Modern synthesis pipelines, spanning generative adversarial networks, diffusion models, neural rendering, and identity-swap architectures, can produce facial imagery whose authenticity is difficult to assess by perceptual inspection alone~\cite{tolosana2020deepfakes,mirsky2021creation}. Recent surveys frame synthetic-media generation and detection as a continuous arms race in which detector reliability depends not only on clean-data accuracy but on robust generalization against distribution shift, compression, post-processing, and adversarial manipulation~\cite{pei2026deepfake,liu2024multimodaldeepfake}. These vulnerabilities threaten misinformation mitigation, biometric authentication, digital-evidence verification, and the dependable deployment of automated forensic systems \cite{radanliev2026threats}.

Detectors based on Convolutional Neural Networks (CNNs)~\cite{he2016deep,huang2017densely,tan2019efficientnet}, Vision Transformers (ViTs)~\cite{dosovitskiy2021image,liu2021swin}, and pretrained face-verification models~\cite{ng2024can} perform strongly on standardized benchmarks such as FaceForensics++~\cite{rossler2019faceforensicspp}. High clean-data accuracy is, however, insufficient for security-critical deployment. From a reliability-engineering standpoint, adversarial examples are systematic, input-dependent failure modes that compromise the dependability of forensic classifiers~\cite{szegedy2014intriguing,goodfellow2015advexamples}, and in deepfake forensics they let manipulated media bypass automated screening~\cite{akhtar2018threat,zhao2024revisiting}. A realistic reliability assessment therefore requires \emph{black-box transfer} evaluation, in which adversarial examples are optimized only on accessible surrogate models and evaluated against blind target detectors whose architectures and parameters are concealed from the attacker.

\textbf{Limitations of existing attack paradigms.}
Transfer-based attacks such as MI-FGSM~\cite{dong2018boosting}, DI-FGSM~\cite{xie2019improving}, and TI-FGSM~\cite{dong2019evading} improve cross-model generalization by stabilizing gradient trajectories or applying stochastic input transformations, but their success rates fall sharply when surrogate and target detectors differ in architecture or attention mechanism~\cite{papernot2017practical}. Query-based attacks~\cite{andriushchenko2020square} rely on extensive target feedback that is unavailable under no-query threat models. Parameter-free ensembles such as AutoAttack~\cite{croce2020reliable} establish strong baselines but lack semantic, image-specific orchestration of heterogeneous perturbations, and reinforcement-learning agents~\cite{domico2025adversarial} add adaptive search without multimodal visual understanding or closed-loop hyperparameter reasoning grounded in forensic feature structure.

\textbf{The deepfake-specific transfer challenge.}
Transferability is uniquely difficult in deepfake detection, because forensic classifiers do not evaluate stable object-level categories but instead exploit weak, localized, generator-dependent traces such as blending boundaries, high-frequency texture inconsistencies, and spatial irregularities~\cite{wang2020cnn,qian2020thinking}. A perturbation that disrupts local convolutional filters in a CNN surrogate may fail to transfer to a ViT target that relies on global token interactions~\cite{naseer2021intriguing,wei2022towards}. Robust transfer therefore requires perturbation diversity across complementary representation spaces, which favors a synthesis of spatial transformations, saliency-driven modifications, frequency-domain manipulations, and block-structured attacks over pixel-level loss maximization alone. Surrogate overfitting is mitigated by the Spectrum Simulation Attack (SSA)~\cite{long2022ssa} through frequency-domain augmentation, while cross-architecture transfer is improved by Block Shuffle and Rotation (BSR)~\cite{wang2024bsr} through block-wise transformations during gradient extraction. Although Vision-Language Models (VLMs) are not trained as forensic classifiers, structured descriptions of facial topology, visible inconsistencies, and compression artifacts can be produced by them, making their semantic guidance a useful complement to numerical surrogate feedback for image-specific attack planning.

\textbf{Remaining gaps in agentic attacks.}
The closest prior approach is the agentic attack ARMOR~\cite{lee2026armor}, which orchestrates three spatial primitives (CW-style, JSMA, STA) through VLM analysis and LLM guidance~\cite{acharya2025,tran2025macollab,abou2025agentic}. Four limitations motivate the present work. \emph{First}, its perturbation space is spatial-only and lacks the frequency-domain and block-structured primitives needed to circumvent ViT-based detectors. \emph{Second}, its mixing score combines a surrogate term with a black-box probability term $p_{\mathrm{bb}}(t\mid\cdot)$, which weakens the no-query property. \emph{Third}, its surrogate ensemble uses only two CNNs (ResNet-50, DenseNet-121), giving insufficient inductive-bias diversity. \emph{Fourth}, evaluation was restricted to a single low-quality (LQ) subset of AADD and a single blind target (ViT-B/16), leaving open whether agentic orchestration is viable against high-quality (HQ) imagery and whether it generalizes to other transformer detectors.

\textbf{Proposed framework.}
In this work, ARMOR++ is proposed as a vision-language-guided agentic framework for method orchestration and reparameterization. Its core mechanism is a closed-loop integration of VLM-derived semantic analysis with LLM-guided primitive selection, adaptive reparameterization, and entropy-regularized mixing, executed under a \emph{strict no-query protocol} that relies on surrogate feedback alone. Qwen2.5-VL~\cite{qwen2025vl} and Qwen3~\cite{yang2025qwen3} are coupled within a multi-agent system where candidates are planned, executed, critiqued, and synthesized by specialized agents. A soft spatial prior is formed from visual anomalies by an \emph{Analysis Agent}, global constraints are set by a \emph{Conductor Agent}, and the five complementary primitives are executed by parallel \emph{Method Agents}
\begin{equation*}
\mathcal{M} = \{\mathrm{CW},\,\mathrm{JSMA},\,\mathrm{STA},\,\mathrm{SSA},\,\mathrm{BSR}\},
\end{equation*}
spanning optimization-, saliency-, spatial-transformation-, frequency-domain-, and block-structured perturbations~\cite{carlini2017towards,papernot2016limitations,xiao2018spatially,long2022ssa,wang2024bsr}. \emph{Advisor} and \emph{Strategist} agents reparameterize hyperparameter bounds, and a \emph{Mixer Agent} selects convex perturbation combinations. No-query blind-target transfer is reported under a matched perturbation envelope, with primary comparisons against the strongest agentic baseline (ARMOR) and the strongest non-agentic baseline (AutoAttack-PGD). Component-removal ablations isolate semantic analysis, reparameterization, stagnation handling, entropy-regularized mixing, SSA, BSR, and the third surrogate. All claims are bounded to the evaluated baselines and ablations.

The main contributions are as follows.
\begin{itemize}
    \item \textbf{Transferable agentic attack framework.} ARMOR++ is a multimodal, multi-agent architecture for no-query evasion of deepfake detectors. All optimization, selection, and reparameterization rely solely on surrogate feedback and LLM-guided reasoning, and the blind target is queried only once, after the adversarial example has been finalized, never to update or re-select a candidate.
    \item \textbf{Semantically guided closed-loop orchestration.} Attack generation is formulated as a closed loop that combines VLM-derived spatial priors with LLM-driven stagnation detection, bounded constraint escalation, adaptive reparameterization, and entropy-regularized mixing.
    \item \textbf{Multi-domain perturbation space.} Five complementary mechanisms integrate frequency-domain (SSA) and block-structured (BSR) primitives alongside optimization-, saliency-, and spatial-transformation attacks, targeting the spectral and patch-based inductive biases of HQ deepfakes and ViT/Swin detectors.
    \item \textbf{Reliability-oriented evaluation.} All ASR values carry $95\%$ Wilson confidence intervals, and pairwise improvements are assessed by exact McNemar tests with Holm--Bonferroni correction. Per-image surrogate forward-pass counts (Table~\ref{tab:compute}), two defenses, and a failure-mode analysis are reported in reliability-engineering terms.
    \item \textbf{Cross-regime evaluation.} Transferability is benchmarked on AADD-2025~\cite{battiato2025aadd} with three CNN surrogates and two transformer blind targets under both LQ and HQ regimes, together with a zero-shot check on DFDC-Preview~\cite{dolhansky2019dfdc}.
\end{itemize}
Compared with ARMOR~\cite{lee2026armor}, ARMOR++ adds SSA and BSR, a third surrogate (EfficientNet-B4), strict no-query mixing without $p_{\mathrm{bb}}$, entropy-regularized candidate synthesis, LQ/HQ evaluation on ViT-B/16 and Swin-B, DFDC-Preview transfer, two defenses, compute accounting, statistical testing, and a failure-mode taxonomy. The inherited agentic control loop is retained, while the new evidence concerns the strict no-query extension, the larger multi-domain primitive set, and the harder cross-regime evaluation.

\section{Related Work}\label{sec:related} 


\subsection{Deepfake and AI-Generated Image Detection}
Deepfake detection has matured from supervised binary classification on curated face-swap datasets to forensic analysis of generator-specific artifacts. FaceForensics++~\cite{rossler2019faceforensicspp} catalyzed spatial detectors, while later designs targeted localized and generalizable evidence: multi-attentional networks~\cite{zhao2021multi} isolate manipulation regions through attention, and Self-Blended Images~\cite{shiohara2022detecting} improve cross-manipulation generalization by learning blending boundaries from pseudo-forgeries. As synthetic media moved toward fully AI-generated scenes, DeepfakeBench~\cite{yan2023deepfakebench} standardized evaluation across generation topologies and zero-shot methods~\cite{cozzolino2024zeroshot} decoupled forensic analysis from specific training distributions. Surveys~\cite{tolosana2020deepfakes,liu2024multimodaldeepfake,pei2026deepfake} confirm that contemporary detectors fuse spatial, spectral, and semantic cues, yet prioritize classification generalization over adversarial reliability, which leaves them vulnerable to systematic perturbations and transfer attacks.

\subsection{Adversarial Vulnerability of Deepfake Detectors}
Carlini and Farid~\cite{carlini2020evading} showed that synthetic-image detectors can be subverted by imperceptible gradient-based attacks and latent-space manipulations. Later work~\cite{gandhi2020adversarial,fernandes2020adversarial} demonstrated failure under $\ell_p$-bounded attacks and localized texture patches, and 2D-Malafide~\cite{galdi2024malafide} showed that lightweight convolutional filtering is an effective non-additive attack vector. The trade-off between evasion and perceptual fidelity was formalized in the AADD-2025 challenge~\cite{battiato2025aadd}, where methods such as MIG-COW~\cite{seo2025migcow} and MS-GAGA~\cite{msgaga2025} combine gradient decomposition with metric-aware candidate selection. These approaches enhance evasion under SSIM constraints but operate as static pipelines, fixing an optimization strategy \emph{a priori} without semantic analysis, dynamic primitive switching, or closed-loop reparameterization. ARMOR++ recasts this static optimization as an agentic orchestration problem.

\subsection{Transferable and Query-Efficient Black-Box Image Attacks}
Black-box attackers rely predominantly on transferability~\cite{liu2017delving,papernot2017practical}. Iterative gradient stabilization~\cite{dong2018boosting}, stochastic input transformations~\cite{xie2019improving,dong2019evading}, and Nesterov-accelerated scale invariance~\cite{Lin2020Nesterov} form the standard arsenal. Query-based attacks~\cite{guo2019simple,andriushchenko2020square,cheng2024pbo} optimize through model feedback but require hundreds to thousands of target queries, violating the no-query model prioritized in security-critical forensics. Recent work emphasizes representation diversity: feature-importance attacks~\cite{wang2021feature} perturb intermediate layers, SSA~\cite{long2022ssa} targets frequency-domain vulnerabilities, and BSR~\cite{wang2024bsr} randomizes block-level gradients to target patch-based ViTs~\cite{wei2022towards}. This multi-domain insight is operationalized in ARMOR++ by deploying geometric, spectral, and block-structured primitives concurrently under an adaptive mixer.

\subsection{Vision-Language Models, LLM Agents, and Automated Attack Design}
Multimodal LLMs have catalyzed two adversarial trajectories, namely attacking the models themselves and using them as agents against downstream systems. AnyAttack~\cite{zhang2025anyattack}, SAAET~\cite{jia2025saaet}, and FOA-Attack~\cite{jia2025foa} show that multimodal LLMs are susceptible to semantic feature misalignment and confirm that semantic alignment governs transferability, although they treat the model as \emph{target}. In parallel, LLMs serve as planning and optimization agents~\cite{wu2024autogen,park2023generative,xi2025rise}: OPRO~\cite{yang2024opro} and AgentHPO~\cite{liu2025agenthpo} refine hyperparameters through natural-language reflection, AutoDA~\cite{fu2022autoda} and L-AutoDA~\cite{guo2024lautoda} discover decision-based attack algorithms through evolutionary loops, and He~\textit{et al.}~\cite{he2025redteaming} study communication-level attacks on multi-agent LLM systems. Current deepfake attacks lack semantic context, and existing LLM-based adversarial agents mostly design static algorithms offline. In contrast, a VLM is used in ARMOR++ as a real-time semantic perception module, and an LLM is used as a hyperparameter optimizer for closed-loop orchestration, critique, and primitive mixing.

\section{Preliminaries}\label{sec:preliminaries}

\subsection{Notation and Threat Model}\label{subsec:threat_model}

Let $\pmb{x} \in \mathcal{X} \subseteq [0,1]^d$ denote an input image, where $d = H \times W \times C$. Let $y \in \{1,\ldots,K\}$ be the ground-truth label and $t \in \{1,\ldots,K\}$ the adversarial target. In deepfake detection $K=2$ with classes \emph{real} and \emph{fake}, and the evasion objective fixes $t = \textit{real}$. A classifier $f : \mathcal{X} \rightarrow \{1,\ldots,K\}$ produces logits $Z_f(\pmb{x}) \in \mathbb{R}^{K}$ and class probabilities
\begin{equation}
  p_f(k \mid \pmb{x}) = \frac{\exp\bigl(Z_{f,k}(\pmb{x})\bigr)}{\sum_{j=1}^{K}\exp\bigl(Z_{f,j}(\pmb{x})\bigr)}.
  \label{eq:softmax}
\end{equation}
The targeted cross-entropy loss for class $t$ is $\mathcal{L}_{\mathrm{CE}}(f, \pmb{x}, t) = -\log p_f(t \mid \pmb{x})$.

\textbf{Threat model.}
This work adopts a \emph{strict no-query black-box transfer} setting. The adversary has white-box access to a surrogate ensemble of $R$ detectors,
\begin{equation}
  \mathcal{F}_{\mathrm{surr}} = \{f_1, \ldots, f_R\},
  \label{eq:surrogate_set}
\end{equation}
but no knowledge of the architecture, parameters, or gradients of the blind target detector $g$, and \emph{never queries} $g$ during optimization, candidate selection, mixing, hyperparameter adaptation, or final selection. The target is evaluated only once, after the final adversarial example has been chosen, and that query is used solely for reporting and is never fed back. This protocol is strictly stronger than prior agentic attacks~\cite{lee2026armor} that permit a black-box term $p_{\mathrm{bb}}(t\mid\cdot)$ in the mixing score. In Section~\ref{sec:experiments}, $R=3$ (ResNet-50, DenseNet-121, EfficientNet-B4), and two blind targets (ViT-B/16, Swin-B) are evaluated.

The weighted surrogate-ensemble probability for the target class is
\begin{equation}
  \bar{p}_{\mathrm{surr}}(t \mid \pmb{x}) = \sum_{r=1}^{R} \alpha_r\, p_{f_r}(t \mid \pmb{x}),\quad \pmb{\alpha} \in \Delta^R,
  \label{eq:weighted_surrogate_probability}
\end{equation}
where $\Delta^R = \{\pmb{\alpha} \in \mathbb{R}_{+}^R : \sum_{r=1}^{R}\alpha_r = 1\}$ and, unless stated otherwise, $\alpha_r = 1/R$. The ensemble targeted loss is
\begin{equation}
  \mathcal{L}_{\mathrm{surr}}(\pmb{x}, t) = \sum_{r=1}^{R} \alpha_r\, \mathcal{L}_{\mathrm{CE}}(f_r, \pmb{x}, t)
  = -\sum_{r=1}^{R} \alpha_r\,\log p_{f_r}(t \mid \pmb{x}).
  \label{eq:surrogate_loss}
\end{equation}
By Jensen's inequality $\mathcal{L}_{\mathrm{surr}}(\pmb{x},t) \ne -\log \bar{p}_{\mathrm{surr}}(t\mid\pmb{x})$ in general, so $\mathcal{L}_{\mathrm{surr}}$ is used for gradient-based optimization and $\bar{p}_{\mathrm{surr}}$ for selection scoring. The objective is to synthesize $\pmb{x}_{\mathrm{adv}} = \pmb{x} + \pmb{\delta}$ that maximizes $\bar{p}_{\mathrm{surr}}(t \mid \pmb{x}_{\mathrm{adv}})$ while perceptual fidelity is preserved, under the expectation that the perturbation transfers to $g$.

\subsection{Perceptual and Perturbation Constraints}\label{subsec:constraints}

The pixel- and perturbation-domain projection operators are $\Pi_{[0,1]}(\pmb{z}) = \operatorname{clip}(\pmb{z}, 0, 1)$ and $\Pi_{\infty,\epsilon}(\pmb{\delta}) = \operatorname{clip}(\pmb{\delta}, {-\epsilon}, \epsilon)$, where $\epsilon > 0$ is the $\ell_{\infty}$ budget. The combined projected-candidate operator is
\begin{equation}
  \mathcal{P}_{\epsilon}(\pmb{x}, \pmb{\delta}) = \Pi_{[0,1]}\!\left(\pmb{x} + \Pi_{\infty,\epsilon}(\pmb{\delta})\right),
  \label{eq:projected_candidate}
\end{equation}
applied by every primitive and the mixer to obtain a valid adversarial image. The feasible adversarial set is
\begin{equation}
  \mathcal{C}(\pmb{x}, \epsilon, \tau) = \bigl\{\, \pmb{z} \in [0,1]^d : \|\pmb{z} - \pmb{x}\|_{\infty} \le \epsilon,\, \mathrm{SSIM}(\pmb{x}, \pmb{z}) \ge \tau \bigr\},
  \label{eq:feasible_set}
\end{equation}
where $\mathrm{SSIM}(\cdot,\cdot) \in [0,1]$~\cite{wang2004ssim} and $\tau \in [0,1]$ is the minimum perceptual-quality threshold set by the Conductor Agent (Section~\ref{subsec:phase1}). Because SSIM is nonlinear, membership in $\mathcal{C}$ is not enforced by projection but enters as a continuous penalty in the mixing score (Section~\ref{subsec:phase4}). The strict transfer objective at outer iteration $k$ is
\begin{equation}
  \begin{aligned}
    \max_{\pmb{\delta}^{(k)}} \quad
    & \bar{p}_{\mathrm{surr}}\!\left(t \,\middle|\, \mathcal{P}_{\epsilon^{(k)}}(\pmb{x}, \pmb{\delta}^{(k)})\right) \\
    \mathrm{s.t.} \quad
    & \mathrm{SSIM}\!\left(\pmb{x},\, \mathcal{P}_{\epsilon^{(k)}}(\pmb{x}, \pmb{\delta}^{(k)})\right) \ge \tau^{(k)},
  \end{aligned}
  \label{eq:strict_transfer_objective}
\end{equation}
where $\epsilon^{(k)}$ and $\tau^{(k)}$ may be updated by the Strategist Agent (Section~\ref{subsec:phase3}).

\begin{figure*}[t]
    \centering
    \includegraphics[width=\linewidth]{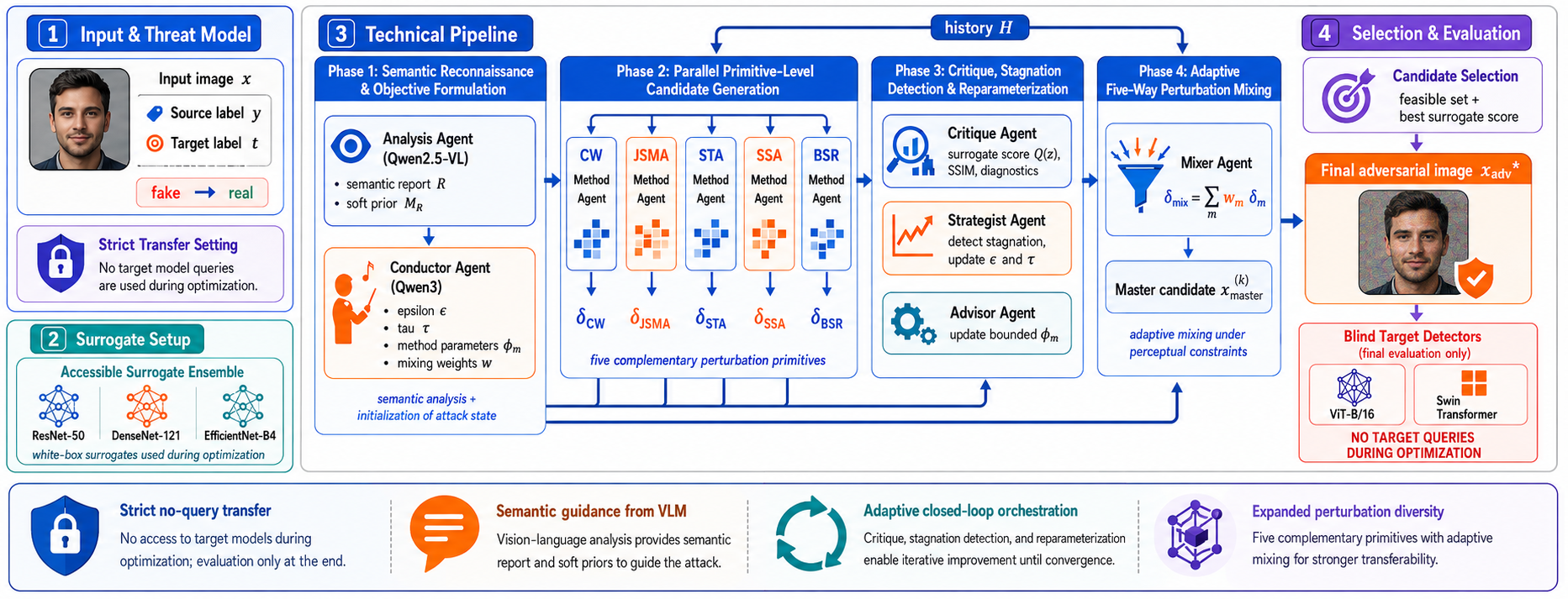}
    \caption{ARMOR++ pipeline. Semantic analysis initializes the attack state, five Method Agents generate candidates, and deterministic critique, strategy, and mixing stages select the master candidate. The blind target $g$ is never queried during execution.}
    \label{fig:armorpp_flowchart}
\end{figure*}

\subsection{Semantic Guidance Representation}\label{subsec:semantic_guidance}

Let $\mathcal{R} = \mathcal{V}(\pmb{x})$ denote the semantic report generated by VLM $\mathcal{V}$ for image $\pmb{x}$, identifying facial topologies, localized texture irregularities, compression artifacts, and candidate regions for perturbation prioritization. The spatial component is a soft prior $\pmb{M}_{\mathcal{R}} \in [0,1]^{H \times W}$, normalized with lower bound $\xi \ll 1$ so that no location is fully suppressed:
\begin{equation}
  \widetilde{\pmb{M}}_{\mathcal{R}} = \frac{\pmb{M}_{\mathcal{R}} + \xi}{\|\pmb{M}_{\mathcal{R}} + \xi\|_{\infty}},
  \label{eq:mask_normalization}
\end{equation}
where the $\ell_{\infty}$ norm is over all spatial locations, so the peak weight equals $1$. Mask values are bounded below by $\xi/(\|\pmb{M}_{\mathcal{R}}\|_{\infty}+\xi)>0$ and above by $1$, and the mask is broadcast across all $C$ channels. For gradient-driven primitives the prior is applied as a Hadamard product with the surrogate ensemble gradient, $\pmb{g}_{\mathcal{R}}(\pmb{x}) = \widetilde{\pmb{M}}_{\mathcal{R}} \odot \nabla_{\pmb{x}} \mathcal{L}_{\mathrm{surr}}(\pmb{x}, t)$, which up-weights gradient components at VLM-identified regions. For JSMA the mask enters the saliency criterion (Section~\ref{subsubsec:jsma}), whereas for STA, SSA, and BSR it acts as a soft spatial weighting prior. Here $\widetilde{M}_i$ denotes the mask value at the spatial location of unrolled index $i$, with the channel coordinate ignored.

\subsection{Adversarial Primitive Family}\label{subsec:primitives}

Five complementary primitives are orchestrated by ARMOR++, $\mathcal{M} = \{\mathrm{CW}, \mathrm{JSMA}, \mathrm{STA}, \mathrm{SSA}, \mathrm{BSR}\}$. CW-style, JSMA, and STA cover optimization-, saliency-, and spatial-transformation perturbations, whereas SSA and BSR extend coverage to the frequency domain and to block-structured transformations. Each primitive $m \in \mathcal{M}$ maps inputs to a candidate perturbation,
\begin{equation}
  \mathcal{A}_m : \bigl(\pmb{x},\, t,\, \mathcal{F}_{\mathrm{surr}},\, \epsilon,\, \tau,\, \pmb{\phi}_m,\, \widetilde{\pmb{M}}_{\mathcal{R}}\bigr) \mapsto \pmb{\delta}_m,
  \label{eq:primitive_map}
\end{equation}
where $\pmb{\phi}_m$ is the hyperparameter vector adapted by the Advisor Agent (Section~\ref{subsec:phase3}).

\subsubsection{CW-style Primitive}\label{subsubsec:cw}
Inspired by the dense perturbation structure of Carlini--Wagner~\cite{carlini2017towards} but replacing the margin loss with the surrogate cross-entropy of Eq.~\eqref{eq:surrogate_loss}, the CW-style $\ell_2$-penalized primitive minimizes $J_{\mathrm{CW}}(\pmb{\delta}) = \|\pmb{\delta}\|_2^2 + c\, \mathcal{L}_{\mathrm{surr}}(\mathcal{P}_{\epsilon}(\pmb{x}, \pmb{\delta}), t)$ with $c > 0$, through the semantic-guided update
\begin{equation}
  \pmb{\delta}^{(u+1)} = \Pi_{\infty,\epsilon}\!\left(\pmb{\delta}^{(u)} - \eta_{\mathrm{CW}}\, \widetilde{\pmb{M}}_{\mathcal{R}} \odot \nabla_{\pmb{\delta}} J_{\mathrm{CW}}(\pmb{\delta}^{(u)})\right).
  \label{eq:cw_update}
\end{equation}
The hyperparameter vector is $\pmb{\phi}_{\mathrm{CW}} = (c, \eta_{\mathrm{CW}}, U_{\mathrm{CW}})$, where $U_{\mathrm{CW}}$ is the number of inner iterations.

\subsubsection{JSMA Primitive (Sparse Saliency Manipulation)}\label{subsubsec:jsma}
The JSMA primitive~\cite{papernot2016limitations} perturbs a small subset of pixels indexed by scalar $i$ through row-major unrolling over $H\times W \times C$. The ensemble-averaged logits are $\bar{Z}_t(\pmb{x}) = \sum_{r}\alpha_r Z_{f_r,t}(\pmb{x})$ and $\bar{Z}_j(\pmb{x}) = \sum_{r}\alpha_r Z_{f_r,j}(\pmb{x})$ for $j \neq t$, and the target and non-target sensitivities are
\begin{equation}
  a_i = \frac{\partial \bar{Z}_t(\pmb{x})}{\partial x_i},\qquad b_i = \sum_{j \neq t} \frac{\partial \bar{Z}_j(\pmb{x})}{\partial x_i}.
  \label{eq:jsma_sensitivities}
\end{equation}
For a pixel pair $(p,q)$ with composite sensitivities $s^{+}_{pq} = a_p + a_q$ and $s^{-}_{pq} = b_p + b_q$, the VLM-reweighted saliency score is
\begin{equation}
  S_{\mathrm{JSMA}}(p, q) = \widetilde{M}_{p}\, \widetilde{M}_{q} \bigl(-\,s^{+}_{pq}\,s^{-}_{pq}\bigr),
  \label{eq:jsma_score}
\end{equation}
subject to $s^{+}_{pq} > 0$ and $s^{-}_{pq} < 0$, the ``$+$'' branch of the saliency map. Only this branch is used, and pixel values are incremented toward the target class, mirroring Papernot \emph{et al.}~\cite{papernot2016limitations}. The highest-scoring feasible pair $(p^{\star}, q^{\star}) = \argmax_{(p,q)} S_{\mathrm{JSMA}}(p,q)$ is updated as
\begin{equation}
  \delta_i^{(u+1)} = \Pi_{\infty,\epsilon}\!\left(\delta_i^{(u)} + \eta_{\mathrm{JSMA}} \cdot \mathbbm{1}\bigl[i \in \{p^{\star}, q^{\star}\}\bigr]\right),
  \label{eq:jsma_update}
\end{equation}
with hyperparameter vector $\pmb{\phi}_{\mathrm{JSMA}} = (\eta_{\mathrm{JSMA}}, B_{\mathrm{JSMA}}, U_{\mathrm{JSMA}})$, where $B_{\mathrm{JSMA}}$ is the number of pixel pairs perturbed per inner iteration, so up to $B_{\mathrm{JSMA}}\,U_{\mathrm{JSMA}}$ pairs are modified over the full run.

\subsubsection{STA Primitive (Geometric Deformation)}\label{subsubsec:sta}
Geometric-consistency vulnerabilities are exploited by the STA primitive~\cite{xiao2018spatially} through smooth spatial deformation. Let $\pmb{u} \in \mathbb{R}^{H \times W \times 2}$ be a dense displacement field and $\mathcal{W}(\pmb{x},\pmb{u})$ a differentiable bilinear warp. STA solves
\begin{equation}
  \min_{\pmb{u}} \quad \mathcal{L}_{\mathrm{surr}}\!\bigl(\mathcal{W}(\pmb{x},\pmb{u}),\, t\bigr) + \theta_{\mathrm{STA}}\,\mathcal{L}_{\mathrm{flow}}(\pmb{u}),
  \label{eq:sta_objective}
\end{equation}
where $\theta_{\mathrm{STA}} > 0$ controls the adversarial--smoothness trade-off and
\begin{equation}
  \mathcal{L}_{\mathrm{flow}}(\pmb{u}) = \sum_{i=1}^{HW}\,\sum_{j \in \mathcal{N}(i)} \bigl\| \pmb{u}_i - \pmb{u}_j \bigr\|_2
  \label{eq:sta_flow}
\end{equation}
is an anisotropic Total Variation regularizer over the four-neighborhood $\mathcal{N}(i)$ of the pixel at unrolled index $i$, following~\cite{xiao2018spatially}. The displacement field is optimized by projected gradient descent,
\begin{equation}
\begin{aligned}
\pmb{u}^{(r+1)}
&= \Pi_{s_{\max}}\!\Bigl(
\pmb{u}^{(r)}
- \eta_{\mathrm{STA}}
\nabla_{\pmb{u}}\,
\mathcal{J}_{\mathrm{STA}}\bigl(\pmb{u}^{(r)}\bigr)
\Bigr), \\
\mathcal{J}_{\mathrm{STA}}(\pmb{u})
&=
\mathcal{L}_{\mathrm{surr}}\!\bigl(
\mathcal{W}(\pmb{x},\pmb{u}),t
\bigr)
+
\theta_{\mathrm{STA}}
\mathcal{L}_{\mathrm{flow}}(\pmb{u}).
\end{aligned}
\label{eq:sta_update}
\end{equation}
where $\Pi_{s_{\max}}$ clips the per-pixel displacement to $\|\pmb{u}_i\|_{\infty} \le s_{\max}$. The additive perturbation $\pmb{\delta}_{\mathrm{STA}} = \mathcal{W}(\pmb{x},\pmb{u}) - \pmb{x}$ is projected by $\mathcal{P}_{\epsilon}$ to enforce the global $\ell_{\infty}$ budget. This projection can degrade the warp when $\|\mathcal{W}(\pmb{x},\pmb{u})-\pmb{x}\|_{\infty}$ exceeds $\epsilon$, but it is retained to enforce the $\ell_{\infty}$ budget. The hyperparameter vector is $\pmb{\phi}_{\mathrm{STA}} = (\eta_{\mathrm{STA}}, \theta_{\mathrm{STA}}, U_{\mathrm{STA}}, s_{\max})$, where $s_{\max}$ bounds the per-pixel displacement.

\subsubsection{SSA Primitive (Spectral Frequency Simulation)}\label{subsubsec:ssa}
Frequency-domain overfitting is reduced by the SSA primitive~\cite{long2022ssa} through optimization over augmented spectra. With $\mathcal{D}$ and $\mathcal{D}^{-1}$ the 2D DCT and its inverse, and random modulation $\omega \sim \Omega_{\mathrm{SSA}}$,
\begin{equation}
  \mathcal{T}_{\omega}^{\mathrm{SSA}}(\pmb{z}) = \mathcal{D}^{-1}\!\left(\pmb{A}_{\omega} \odot \mathcal{D}(\pmb{z})\right),
  \label{eq:ssa_transform}
\end{equation}
where $\pmb{A}_{\omega} = \pmb{1} + \pmb{\sigma}_{\omega}$ with $\pmb{\sigma}_{\omega} \sim \mathcal{N}(\pmb{0}, \rho_{\mathrm{SSA}}^2 \pmb{I})$ truncated to $[-1+\delta_{\min},\,\infty)$ with $\delta_{\min}=0.05$, ensuring $\pmb{A}_{\omega}\ge \delta_{\min}>0$ and preserving coefficient signs. This truncation preserves the spectral-simulation semantics of the original SSA. The objective is the Monte Carlo expectation over $Q_{\mathrm{SSA}}$ samples,
\begin{equation}
  \widehat{J}_{\mathrm{SSA}}(\pmb{\delta}) = \frac{1}{Q_{\mathrm{SSA}}} \sum_{q=1}^{Q_{\mathrm{SSA}}} \mathcal{L}_{\mathrm{surr}}\!\left(\mathcal{T}_{\omega_q}^{\mathrm{SSA}}\!\bigl(\mathcal{P}_{\epsilon}(\pmb{x}, \pmb{\delta})\bigr),\, t \right),
  \label{eq:ssa_mc}
\end{equation}
with $\omega_q \overset{\mathrm{i.i.d.}}{\sim} \Omega_{\mathrm{SSA}}$, minimized by sign-based descent
\begin{equation}
  \pmb{\delta}^{(u+1)} = \Pi_{\infty,\epsilon}\!\left(\pmb{\delta}^{(u)} - \eta_{\mathrm{SSA}}\, \operatorname{sign}\!\left(\nabla_{\pmb{\delta}}\,\widehat{J}_{\mathrm{SSA}}(\pmb{\delta}^{(u)})\right)\right).
  \label{eq:ssa_update}
\end{equation}
The hyperparameter vector is $\pmb{\phi}_{\mathrm{SSA}} = (\eta_{\mathrm{SSA}}, U_{\mathrm{SSA}}, Q_{\mathrm{SSA}}, \rho_{\mathrm{SSA}})$.

\subsubsection{BSR Primitive (Block-Structured Augmentation)}\label{subsubsec:bsr}
Transfer to patch-based ViTs is improved by the BSR primitive~\cite{wang2024bsr} through disruption of globally coherent structure. The image is partitioned into $N_b$ non-overlapping blocks $\{\pmb{B}_j\}_{j=1}^{N_b}$. A random transformation $\nu \sim \Omega_{\mathrm{BSR}}$ specifies a permutation $\pi$ drawn from the neighborhood subset $\mathcal{S}_{N_b}^{\mathrm{nbhd}} \subseteq \mathcal{S}_{N_b}$ that swaps only blocks within $\ell_1$-distance $1$ on the block grid, following~\cite{wang2024bsr}, together with an i.i.d.\ rotation $r_j \in \{0^{\circ},90^{\circ},180^{\circ},270^{\circ}\}$ per block:
\begin{equation}
  \mathcal{T}_{\nu}^{\mathrm{BSR}}(\pmb{z}) = \operatorname{Merge}\!\left(\left\{\operatorname{Rot}_{r_j}\!\bigl(\pmb{B}_{\pi(j)}\bigr)\right\}_{j=1}^{N_b}\right),
  \label{eq:bsr_transform}
\end{equation}
where $\operatorname{Merge}(\cdot)$ reassembles the blocks. Restricting to $\mathcal{S}_{N_b}^{\mathrm{nbhd}}$ preserves coarse semantic plausibility while inducing block-level gradient diversity. The objective is
\begin{equation}
  \widehat{J}_{\mathrm{BSR}}(\pmb{\delta}) = \frac{1}{Q_{\mathrm{BSR}}} \sum_{q=1}^{Q_{\mathrm{BSR}}} \mathcal{L}_{\mathrm{surr}}\!\left(\mathcal{T}_{\nu_q}^{\mathrm{BSR}}\!\bigl(\mathcal{P}_{\epsilon}(\pmb{x}, \pmb{\delta})\bigr),\, t \right),
  \label{eq:bsr_mc}
\end{equation}
updated by sign-based descent as in SSA, with hyperparameter vector $\pmb{\phi}_{\mathrm{BSR}} = (\eta_{\mathrm{BSR}}, U_{\mathrm{BSR}}, Q_{\mathrm{BSR}}, N_b)$.

\section{Proposed Methodology}\label{sec:methodology}

\begin{algorithm}[!t]
\scriptsize
\caption{ARMOR++ Algorithm}
\label{alg:armorpp_framework}
\begin{algorithmic}[1]
\Require Clean image $\pmb{x}$ and labels $(y,t)$, surrogate ensemble $\mathcal{F}_{\mathrm{surr}}$, primitive set $\mathcal{M}$, maximum iterations $K_{\max}$, bounds $(\epsilon_{\max},\tau_{\min})$
\Ensure Final adversarial image $\pmb{x}_{\mathrm{adv}}^{\star}$

\Statex \Comment{\textbf{Phase 1: Semantic Reconnaissance and Initialization}}
\State Generate $\mathcal{R}^{(0)}$ via the \textit{Analysis Agent}.
\State Compute $\widetilde{\pmb{M}}_{\mathcal{R}}$ using \eqref{eq:mask_normalization}.
\State Initialize $\mathcal{S}^{(0)}$ via the \textit{Conductor Agent} using \eqref{eq:initial_state}.
\State Initialize $\mathcal{H}^{(0)}\leftarrow\emptyset$ and $\mathcal{Z}\leftarrow\emptyset$.

\For{$k=0$ to $K_{\max}-1$}

    \Statex \Comment{\textbf{Phase 2: Parallel Primitive-Level Generation}}
    \ForAll{$m\in\mathcal{M}$ \textbf{in parallel}}
        \State Compute $\pmb{\delta}_{m}^{(k)}$ via $\mathcal{A}_m$ and $\pmb{x}_{m}^{(k)}\leftarrow \mathcal{P}_{\epsilon^{(k)}}(\pmb{x},\pmb{\delta}_{m}^{(k)})$.
        \State Compute $\pmb{d}_{m}^{(k)}$ using \eqref{eq:method_diagnostics} and archive $\pmb{x}_{m}^{(k)}$ in $\mathcal{Z}$.
    \EndFor

    \Statex \Comment{\textbf{Phase 3: Critique and Next-State Adaptation}}
    \State Form $\mathcal{B}^{(k)}$ and score it via the \textit{Critique Agent} using \eqref{eq:quality_surrogate_score}.
    \If{the \textit{Strategist Agent} detects stagnation using \eqref{eq:stagnation_condition}}
        \State Relax $(\epsilon^{(k+1)},\tau^{(k+1)})$ using \eqref{eq:constraint_updates}.
    \Else
        \State Set $(\epsilon^{(k+1)},\tau^{(k+1)})\leftarrow(\epsilon^{(k)},\tau^{(k)})$.
    \EndIf
    \State Update $\pmb{\phi}_{m}^{(k+1)}$ for all $m$ via the \textit{Advisor Agent} using \eqref{eq:advisor_update}.

    \Statex \Comment{\textbf{Phase 4: Adaptive Five-Way Mixing}}
    \State Solve for $\pmb{w}^{\star(k)}$ via the \textit{Mixer Agent} using \eqref{eq:mixer_score}.
    \State Synthesize $\pmb{x}_{\mathrm{master}}^{(k)}$, archive it in $\mathcal{Z}$, and update $\mathcal{H}^{(k+1)}$.

\EndFor

\Statex \Comment{\textbf{Final Selection: No-Query Target Evaluation}}
\State Select $\pmb{x}_{\mathrm{adv}}^{\star}$ from $\mathcal{Z}$ using \eqref{eq:feasible_indicator}--\eqref{eq:final_selection}.
\State Evaluate $\pmb{x}_{\mathrm{adv}}^{\star}$ on blind target $g$ \emph{for final reporting only}.
\State \Return $\pmb{x}_{\mathrm{adv}}^{\star}$.
\end{algorithmic}
\end{algorithm}

As shown in Fig.~\ref{fig:armorpp_flowchart} and detailed in Algorithm~\ref{alg:armorpp_framework}, static pre-ordered pipelines are replaced by a multi-agent closed loop in ARMOR++, where generation is decomposed into four phases per outer iteration. Three principles govern the framework: the strict no-query protocol requires that all optimization and adaptation rely exclusively on $\mathcal{F}_{\mathrm{surr}}$, multi-domain coverage is obtained by deploying all five primitives in $\mathcal{M}$, and closed-loop orchestration enables autonomous stagnation detection, hyperparameter adaptation, and mixing-weight adjustment.

\noindent\textbf{Language-model decisions.}
The generative components are limited to semantic perception, initialization, and reparameterization: the VLM Analysis Agent emits $\mathcal{R}$ and $\widetilde{\pmb{M}}_{\mathcal{R}}$, the LLM Conductor initializes $\mathcal{S}^{(0)}$, and the LLM Advisor proposes $\Delta\pmb{\phi}_m^{(k)}$ before projection by Eq.~\eqref{eq:advisor_update}. The Critique, Strategist, and Mixer are deterministic modules implementing Eqs.~\eqref{eq:quality_surrogate_score}--\eqref{eq:mixer_score}. These roles are isolated by the ablations in Section~\ref{subsec:ablation}.

At outer iteration $k$, candidates are produced by the Method Agents $\{\pmb{\delta}_{m}^{(k)}\}_{m \in \mathcal{M}}$, and the Mixer synthesizes the master candidate
\begin{equation}
  \pmb{x}_{\mathrm{master}}^{(k)} = \mathcal{P}_{\epsilon^{(k)}}\!\left(\pmb{x},\, \sum_{m\in\mathcal{M}} w_m^{(k)}\, \pmb{\delta}_{m}^{(k)}\right),\quad \pmb{w}^{(k)} \in \Delta^{|\mathcal{M}|}.
  \label{eq:master_candidate_general}
\end{equation}
Each $\|\pmb{\delta}_{m}^{(k)}\|_{\infty} \le \epsilon^{(k)}$, so the convex combination satisfies the budget and only the $\Pi_{[0,1]}$ box clip of $\mathcal{P}_{\epsilon^{(k)}}$ is active.

\subsection{Phase 1: Semantic Reconnaissance and Objective Formulation}\label{subsec:phase1}

A structured report $\mathcal{R}^{(0)} = \mathcal{V}_{\mathrm{Qwen2.5\text{-}VL}}(\pmb{x})$ is produced by the \textit{Analysis Agent} (Qwen2.5-VL-32B-Instruct-AWQ~\cite{qwen2025vl}) to capture facial semantics, localized artifacts, texture properties, and forensic-vulnerability regions. It is converted to $\widetilde{\pmb{M}}_{\mathcal{R}}$ by Eq.~\eqref{eq:mask_normalization} and shared read-only with all Method Agents. A \textit{Conductor Agent} (Qwen3-32B-AWQ~\cite{yang2025qwen3}) is then used to synthesize $\mathcal{R}^{(0)}$, labels $(y,t)$, and baseline surrogate predictions into
\begin{equation}
  \mathcal{S}^{(0)} = \Bigl(\epsilon^{(0)},\, \tau^{(0)},\, \bigl\{\pmb{\phi}_{m}^{(0)}\bigr\}_{m\in\mathcal{M}},\, \pmb{w}^{(0)}\Bigr),
  \label{eq:initial_state}
\end{equation}
subject to $0 < \epsilon^{(0)} \le \epsilon_{\max}$ and $\tau_{\min} \le \tau^{(0)} \le 1$. The initial $\pmb{\phi}_{m}^{(0)}$ are image-adaptive values informed by $\mathcal{R}^{(0)}$, seeded from the default configuration of Section~\ref{subsec:impl}. The global objective is Eq.~\eqref{eq:strict_transfer_objective}.

\subsection{Phase 2: Parallel Primitive-Level Candidate Generation}\label{subsec:phase2}

At iteration $k$, the algorithms in $\mathcal{M}$ are invoked concurrently by five \textit{Method Agents}. 
Each agent processes the state tuple $\Xi^{(k)} = (\pmb{x}, t, \mathcal{F}_{\mathrm{surr}}, \epsilon^{(k)}, \tau^{(k)}, \pmb{\phi}_{m}^{(k)}, \widetilde{\pmb{M}}_{\mathcal{R}}, \mathcal{H}^{(k)})$, 
where $\mathcal{H}^{(k)}$ denotes the shared history. The agents produce $\pmb{\delta}_{m}^{(k)}$ and 
$\pmb{x}_{m}^{(k)} = \mathcal{P}_{\epsilon^{(k)}}(\pmb{x},\pmb{\delta}_{m}^{(k)})$, 
with performance compressed into the diagnostic vector
\begin{equation}
\pmb{d}_{m}^{(k)}
=
\begin{bmatrix}
\bar{p}_{\mathrm{surr}}\!\left(t \mid \pmb{x}_{m}^{(k)}\right) \\
\mathcal{L}_{\mathrm{surr}}\!\left(\pmb{x}_{m}^{(k)},t\right) \\
\mathrm{SSIM}\!\left(\pmb{x},\pmb{x}_{m}^{(k)}\right) \\
\left\|\pmb{\delta}_{m}^{(k)}\right\|_{\infty}
\end{bmatrix}.
\label{eq:method_diagnostics}
\end{equation}
Independent agent execution is maintained, and coordination is deferred to the Mixer in Phase 4.

\subsection{Phase 3: Critique, Stagnation Detection, and Adaptive Reparameterization}\label{subsec:phase3}

The candidate pool is scored by a \textit{Critique Agent}
\begin{equation}
  \mathcal{B}^{(k)} = \begin{cases}
    \bigl\{\pmb{x}_{m}^{(k)} : m\in\mathcal{M}\bigr\}, & k=0, \\[3pt]
    \bigl\{\pmb{x}_{m}^{(k)} : m\in\mathcal{M}\bigr\} \cup \bigl\{\pmb{x}_{\mathrm{master}}^{(k-1)}\bigr\}, & k>0,
  \end{cases}
  \label{eq:candidate_pool}
\end{equation}
where each $\pmb{z} \in \mathcal{B}^{(k)}$ receives
\begin{equation}
\begin{aligned}
\mathcal{Q}^{(k)}(\pmb{z})
&=
\mu_p\,\bar{p}_{\mathrm{surr}}\!\left(t \mid \pmb{z}\right)
-
\mu_{\ell}\,\widetilde{\mathcal{L}}_{\mathrm{surr}}\!\left(\pmb{z},t\right) \\
&\quad
-
\mu_s\,
\left[
\tau^{(k)}
-
\mathrm{SSIM}\!\left(\pmb{x},\pmb{z}\right)
\right]_{+}.
\end{aligned}
\label{eq:quality_surrogate_score}
\end{equation}
with $[a]_{+} = \max(a,0)$ and $\widetilde{\mathcal{L}}_{\mathrm{surr}}(\pmb{z},t) = \mathcal{L}_{\mathrm{surr}}(\pmb{z},t)/\log K$ the ensemble cross-entropy scaled by $1/\log K$ (with $K=2$ classes, distinct from the iteration budget $K_{\max}$), so a uniform-probability prediction maps to $1$. The weights satisfy $\mu_p, \mu_\ell, \mu_s \ge 0$. The pool scores are written to the shared history $\mathcal{H}^{(k)}$ read by the Advisor Agent, whereas only the master score $\mathcal{Q}^{(k)}(\pmb{x}_{\mathrm{master}}^{(k)})$ drives the stagnation test of Eq.~\eqref{eq:stagnation_condition}, and the final selection of Eq.~\eqref{eq:final_selection} ranks by surrogate confidence rather than by $\mathcal{Q}$.

\noindent\textbf{Stagnation detection.}
Master scores are monitored by a \textit{Strategist Agent} over $I_k = \{k-W,\ldots,k-1\} \cap \mathbb{N}_0$, evaluated for $k\ge 1$. Stagnation is declared when
\begin{equation}
  \left(\max_{i\in I_k}\mathcal{Q}^{(i)}\!\left(\pmb{x}_{\mathrm{master}}^{(i)}\right) < \kappa\right) \land \left(\operatorname*{range}_{i\in I_k} \mathcal{Q}^{(i)}\!\left(\pmb{x}_{\mathrm{master}}^{(i)}\right) < \varrho\right),
  \label{eq:stagnation_condition}
\end{equation}
with $\kappa \in (0,\mu_p)$ and $\varrho>0$. Upon stagnation,
\begin{equation}
\begin{aligned}
\epsilon^{(k+1)}
&=
\min\!\left(
\epsilon_{\max},
\epsilon^{(k)}+\Delta_{\epsilon}
\right), \\
\tau^{(k+1)}
&=
\max\!\left(
\tau_{\min},
\tau^{(k)}-\Delta_{\tau}
\right).
\end{aligned}
\label{eq:constraint_updates}
\end{equation}
and otherwise $\epsilon^{(k+1)} \leftarrow \epsilon^{(k)}$ and $\tau^{(k+1)} \leftarrow \tau^{(k)}$. Every ablation and matched-envelope comparison uses the identical $(\Delta_\epsilon,\Delta_\tau,\epsilon_{\max},\tau_{\min})$ schedule, so no configuration has a larger worst-case budget than another. The escalation frequency is reported in the supplementary material.

\noindent\textbf{Adaptive reparameterization.}
The history $\mathcal{H}^{(k)}$ is read by an \textit{Advisor Agent}, which recommends shifts $\{\Delta\pmb{\phi}_{m}^{(k)}\}$ that are projected onto admissible domains $\Phi_m$,
\begin{equation}
  \pmb{\phi}_{m}^{(k+1)} = \Pi_{\Phi_m}\!\left(\pmb{\phi}_{m}^{(k)} + \Delta\pmb{\phi}_{m}^{(k)}\right),
  \label{eq:advisor_update}
\end{equation}
which guarantees that the proposed updates remain numerically valid regardless of the model output.

\subsection{Phase 4: Adaptive Five-Way Perturbation Mixing}\label{subsec:phase4}

The weight vector $\pmb{w}^{(k)} \in \Delta^{5}$ is optimized by the \textit{Mixer Agent}. For a given $\pmb{w}$,
\begin{equation}
  \pmb{x}_{\mathrm{mix}}^{(k)}(\pmb{w}) = \Pi_{[0,1]}\!\left(\pmb{x} + \sum_{m\in\mathcal{M}} w_m\, \pmb{\delta}_{m}^{(k)}\right),
  \label{eq:mix_candidate}
\end{equation}
where the inner $\ell_\infty$ projection is dropped because its argument already satisfies the budget. The optimal weights are $\pmb{w}^{\star(k)} = \argmax_{\pmb{w} \in \Delta^{5}} S^{(k)}(\pmb{w})$, with mixing score
\begin{equation}
\begin{aligned}
S^{(k)}(\pmb{w})
&=
\lambda_1\,
\bar{p}_{\mathrm{surr}}\!\left(
t \mid \pmb{x}_{\mathrm{mix}}^{(k)}(\pmb{w})
\right) \\
&\quad
-
\lambda_2
\left[
\tau^{(k)}
-
\mathrm{SSIM}\!\left(
\pmb{x},
\pmb{x}_{\mathrm{mix}}^{(k)}(\pmb{w})
\right)
\right]_{+} \\
&\quad
-
\lambda_3
\left\|
\pmb{x}_{\mathrm{mix}}^{(k)}(\pmb{w})
-
\pmb{x}
\right\|_2^2
+
\lambda_4^{(k)}
\mathcal{H}_{\Delta}(\pmb{w}).
\end{aligned}
\label{eq:mixer_score}
\end{equation}
where target-class confidence, the SSIM-violation penalty, perturbation energy, and primitive diversity are balanced through the smoothed entropy regularizer $\mathcal{H}_{\Delta}(\pmb{w}) = -\sum_{m\in\mathcal{M}} w_m\log(w_m + \varepsilon_0)$ with $\varepsilon_0 > 0$. Unlike~\cite{lee2026armor}, the score omits any black-box term $p_{\mathrm{bb}}(t \mid \cdot)$, which enforces the strict no-query protocol. The coefficient $\lambda_4^{(k)} = \lambda_4^{(0)} \exp(-\rho_{\mathrm{mix}}\, k)$, $\rho_{\mathrm{mix}} > 0$, is annealed to transition from exploration to exploitation. The argmax is solved by randomized hill climbing over $\Delta^{5}$ to avoid gradients through SSIM, and the master candidate is $\pmb{x}_{\mathrm{master}}^{(k)} = \pmb{x}_{\mathrm{mix}}^{(k)}(\pmb{w}^{\star(k)})$. The history is updated with the diagnostics, weights, master, score, and next-state constraints and hyperparameters.

\subsection{Candidate Acceptance and Final Selection}\label{subsec:final_selection}

Every primitive candidate $\pmb{x}_{m}^{(k)}$ and master $\pmb{x}_{\mathrm{master}}^{(k)}$ are stored in a cumulative archive $\mathcal{Z}$. A feasibility indicator is
\begin{equation}
\begin{aligned}
\Gamma(\pmb{z})
&=
\mathbbm{1}\!\left[\pmb{z}\in[0,1]^d\right] \\
&\quad \cdot
\mathbbm{1}\!\left[
\left\|\pmb{z}-\pmb{x}\right\|_{\infty}
\le
\epsilon_{\max}
\right] \\
&\quad \cdot
\mathbbm{1}\!\left[
\mathrm{SSIM}(\pmb{x},\pmb{z})
\ge
\tau_{\min}
\right].
\end{aligned}
\label{eq:feasible_indicator}
\end{equation}
and the final adversarial example is the feasible candidate with the highest surrogate target confidence,
\begin{equation}
  \pmb{x}_{\mathrm{adv}}^{\star} = \argmax_{\pmb{z} \in \mathcal{Z}}\, \Bigl(\Gamma(\pmb{z})\cdot \bar{p}_{\mathrm{surr}}(t \mid \pmb{z})\Bigr).
  \label{eq:final_selection}
\end{equation}
Within $\epsilon_{\mathrm{tie}}=10^{-4}$ of the maximum, masters are preferred over primitives, and remaining ties are broken by the smaller $\|\pmb{z}-\pmb{x}\|_2$. Only $\pmb{x}_{\mathrm{adv}}^{\star}$ is evaluated against $g$, for reporting only, and no candidate is re-ranked after $g$ is observed. The component ablations use the identical archive and selection rule.

\section{Experimental Setup}\label{sec:experiments}

\subsection{Dataset and Evaluation Regimes}\label{subsec:dataset}

ARMOR++ is evaluated on the AADD-2025 benchmark~\cite{battiato2025aadd}, an ACM Multimedia 2025 grand challenge for adversarial attacks against deepfake detectors under perceptual-quality constraints. The benchmark contains AI-generated facial images from GAN- and diffusion-based generators in two disjoint regimes, AADD-LQ (low-quality) and AADD-HQ (high-quality). The evaluation uses $N=713$ fake images on AADD-LQ and $N=693$ on AADD-HQ. The LQ subset is matched to the partition of~\cite{lee2026armor}, and the HQ subset is stratified so that each generator family appears in the same proportion as in AADD-LQ. Because all evaluated images are fake, the evasion objective is to induce the prediction \emph{real}. A held-out $200$-image subset of DFDC-Preview~\cite{dolhansky2019dfdc} provides a strict zero-shot transfer across both attack and data distribution (Section~\ref{subsec:crossdataset}), with the AADD-2025 surrogates and targets used unchanged. Images are resized to $224\times224$ and normalized with ImageNet statistics, and unless stated otherwise all attacks use $\epsilon=8/255$.

\subsection{Detectors and Clean-Data Baseline}\label{subsec:detectors}

The surrogates are ResNet-50~\cite{he2016deep}, DenseNet-121~\cite{huang2017densely}, and EfficientNet-B4~\cite{tan2019efficientnet}. ResNet-50 and DenseNet-121 contribute residual and dense connectivity, while EfficientNet-B4 diversifies the inductive biases and reduces overfitting to the ResNet/DenseNet family, an effect isolated in the ablation (Section~\ref{subsec:ablation}, row (h)). The blind targets are ViT-B/16~\cite{dosovitskiy2021image} and Swin-B~\cite{liu2021swin}, contrasting global self-attention with hierarchical shifted-window attention. Results are treated as evidence across two architectures rather than as a claim about all transformer families, and no target logits, gradients, or queries are used during generation, reparameterization, mixing, or final selection.

All five detectors were fine-tuned on the AADD-2025 training split (disjoint from the LQ/HQ test subsets) with cross-entropy and AdamW (initial lr $1{\times}10^{-4}$, cosine decay, weight decay $0.05$, batch $64$), for up to $50$ epochs with early stopping on validation AUC (patience $5$). As shown in Table~\ref{tab:clean_acc}, clean accuracy exceeds $96\%$ for every detector in both regimes. Thus, the robustness differences reported later are not attributable to weak clean baselines. All backbones are initialized from public ImageNet-pretrained weights, which constitutes a potential implicit representation channel between surrogates and targets that is revisited in Section~\ref{subsec:limitations}.

\begin{table}[!t]
\centering
\caption{Clean-data accuracy (\%) on AADD-LQ ($N=713$) and AADD-HQ ($N=693$), with $95\%$ Wilson CIs. The three CNNs are surrogates, and ViT-B/16 and Swin-B are blind targets.}
\label{tab:clean_acc}
\begin{adjustbox}{max width=\linewidth}
\begin{tabular}{lcc}
\toprule
\textbf{Detector} & \textbf{AADD-LQ Acc.\ (\%)} & \textbf{AADD-HQ Acc.\ (\%)} \\
\midrule
ResNet-50~\cite{he2016deep}                 & $97.46\,[96.04,98.41]$ & $96.62\,[95.04,97.74]$ \\
DenseNet-121~\cite{huang2017densely}        & $98.17\,[96.93,98.92]$ & $97.18\,[95.71,98.20]$ \\
EfficientNet-B4~\cite{tan2019efficientnet}  & $98.31\,[97.10,99.03]$ & $97.32\,[95.86,98.32]$ \\
\midrule
ViT-B/16~\cite{dosovitskiy2021image}        & $98.59\,[97.43,99.24]$ & $97.46\,[96.04,98.41]$ \\
Swin-B~\cite{liu2021swin}                   & $98.73\,[97.61,99.34]$ & $97.61\,[96.21,98.51]$ \\
\bottomrule
\end{tabular}
\end{adjustbox}
\end{table}

\subsection{Baselines}\label{subsec:baselines}

ARMOR++ is evaluated against four baseline groups. The \emph{transfer-based} group comprises MI-FGSM~\cite{dong2018boosting}, DI-FGSM~\cite{xie2019improving}, TI-FGSM~\cite{dong2019evading}, and SINI-FGSM~\cite{Lin2020Nesterov}. The \emph{query-based} group comprises Square~\cite{andriushchenko2020square}, SimBA-DCT~\cite{guo2019simple}, and PBO~\cite{cheng2024pbo}. To preserve the no-query target protocol, these are not permitted to query the blind targets, and following~\cite{lee2026armor} their budget is spent on the surrogate ensemble and evaluated once on the targets, capped at $2{,}500$ surrogate forward passes per image. The \emph{ensemble-based} group comprises AutoAttack-PGD, AutoAttack-Square~\cite{croce2020reliable}, and MS-GAGA~\cite{msgaga2025}. The \emph{agentic} group comprises RL-PPO~\cite{domico2025adversarial} and ARMOR~\cite{lee2026armor}, with ARMOR evaluated under the same strict no-query mixing score (no $p_{\mathrm{bb}}$ term) in its native two-CNN configuration (ResNet-50, DenseNet-121). All surrogate-based methods other than ARMOR use the same three-detector ensemble and $\epsilon = 8/255$, and Table~\ref{tab:compute} reports the per-image surrogate forward-pass count to address the compute-budget confound.

\renewcommand{\arraystretch}{0.95}
\setlength{\tabcolsep}{3pt}

\begin{table*}[t]
\centering
\caption{Blind-target performance on AADD-LQ ($N=713$, majority over $3$ seeds). ASR, wASR, and SSIM are reported with CIs or mean $\pm$ one standard deviation. Bold marks the best column value.}
\label{tab:lq_blind_main}
\begin{adjustbox}{max width=\linewidth}
\begin{tabular}{llcccccc}
\toprule
\multirow{2}{*}{\textbf{Category}} &
\multirow{2}{*}{\textbf{Method}} &
\multicolumn{3}{c}{\textbf{ViT-B/16 (blind)}} &
\multicolumn{3}{c}{\textbf{Swin-B (blind)}} \\
\cmidrule(lr){3-5}\cmidrule(lr){6-8}
& &
\textbf{ASR [95\% CI]} & \textbf{wASR [95\% CI]} & \textbf{SSIM} &
\textbf{ASR [95\% CI]} & \textbf{wASR [95\% CI]} & \textbf{SSIM} \\
\midrule
Transfer  & MI-FGSM~\cite{dong2018boosting}      & .068\,[.052,.089] & .051\,[.038,.067] & .743\,$\pm$\,.038 & .054\,[.040,.073] & .040\,[.029,.054] & .743\,$\pm$\,.038 \\
Transfer  & DI-FGSM~\cite{xie2019improving}      & .038\,[.026,.055] & .032\,[.022,.048] & .846\,$\pm$\,.033 & .047\,[.034,.065] & .040\,[.028,.057] & .846\,$\pm$\,.033 \\
Transfer  & TI-FGSM~\cite{dong2019evading}       & .151\,[.126,.181] & .136\,[.114,.165] & .904\,$\pm$\,.036 & .138\,[.114,.166] & .125\,[.103,.151] & .904\,$\pm$\,.036 \\
Transfer  & SINI-FGSM~\cite{Lin2020Nesterov}     & .059\,[.044,.079] & .042\,[.030,.057] & .711\,$\pm$\,.039 & .043\,[.030,.060] & .031\,[.021,.045] & .711\,$\pm$\,.039 \\
\midrule
Query     & Square~\cite{andriushchenko2020square}& .008\,[.004,.018] & .007\,[.003,.015] & .881\,$\pm$\,.084 & .006\,[.002,.014] & .005\,[.002,.012] & .881\,$\pm$\,.084 \\
Query     & SimBA-DCT~\cite{guo2019simple}       & .004\,[.001,.012] & .004\,[.001,.011] & .968\,$\pm$\,.031 & .003\,[.001,.010] & .003\,[.001,.009] & .968\,$\pm$\,.031 \\
Query     & PBO~\cite{cheng2024pbo}              & .003\,[.001,.010] & .003\,[.001,.009] & .970\,$\pm$\,.030 & .002\,[.000,.008] & .002\,[.000,.007] & .970\,$\pm$\,.030 \\
\midrule
Ensemble  & AA-PGD~\cite{croce2020reliable}      & .196\,[.168,.228] & .108\,[.087,.132] & .559\,$\pm$\,.038 & .183\,[.156,.214] & .101\,[.082,.124] & .559\,$\pm$\,.038 \\
Ensemble  & AA-Square~\cite{croce2020reliable}   & .076\,[.058,.098] & .047\,[.034,.064] & .716\,$\pm$\,.142 & .061\,[.046,.082] & .038\,[.026,.054] & .716\,$\pm$\,.142 \\
Ensemble  & MS-GAGA~\cite{msgaga2025}            & .061\,[.046,.082] & .052\,[.038,.071] & .758\,$\pm$\,.131 & .049\,[.035,.067] & .037\,[.025,.052] & .758\,$\pm$\,.131 \\
\midrule
Agentic   & RL-PPO~\cite{domico2025adversarial}  & .003\,[.001,.010] & .003\,[.001,.009] & \textbf{.972\,$\pm$\,.030} & .002\,[.000,.008] & .002\,[.000,.007] & \textbf{.972\,$\pm$\,.030} \\
\rowcolor{ARMORColor}
Agentic   & ARMOR~\cite{lee2026armor}    & .396\,[.361,.432] & .280\,[.250,.312] & .698\,$\pm$\,.173 & .371\,[.337,.407] & .261\,[.232,.292] & .695\,$\pm$\,.176 \\
\rowcolor{ARMORPPColor}
Agentic   & \textbf{ARMOR++ (ours)}              & \textbf{.443\,[.407,.480]} & \textbf{.312\,[.281,.345]} & .691\,$\pm$\,.169 & \textbf{.408\,[.373,.444]} & \textbf{.287\,[.257,.319]} & .686\,$\pm$\,.176 \\
\bottomrule
\end{tabular}
\end{adjustbox}
\end{table*}

\begin{table*}[t]
\centering
\caption{Blind-target performance on AADD-HQ ($N=693$, majority over $3$ seeds). ASR, wASR, and SSIM are reported with CIs or mean $\pm$ one standard deviation. Bold marks the best column value.}
\label{tab:hq_blind_main}
\begin{adjustbox}{max width=\linewidth}
\begin{tabular}{llcccccc}
\toprule
\multirow{2}{*}{\textbf{Category}} &
\multirow{2}{*}{\textbf{Method}} &
\multicolumn{3}{c}{\textbf{ViT-B/16 (blind)}} &
\multicolumn{3}{c}{\textbf{Swin-B (blind)}} \\
\cmidrule(lr){3-5}\cmidrule(lr){6-8}
& &
\textbf{ASR [95\% CI]} & \textbf{wASR [95\% CI]} & \textbf{SSIM} &
\textbf{ASR [95\% CI]} & \textbf{wASR [95\% CI]} & \textbf{SSIM} \\
\midrule
Transfer  & MI-FGSM~\cite{dong2018boosting}      & .041\,[.029,.058] & .031\,[.021,.045] & .784\,$\pm$\,.029 & .029\,[.019,.044] & .022\,[.014,.034] & .784\,$\pm$\,.029 \\
Transfer  & DI-FGSM~\cite{xie2019improving}      & .022\,[.013,.036] & .019\,[.011,.032] & .872\,$\pm$\,.026 & .018\,[.010,.031] & .015\,[.008,.026] & .872\,$\pm$\,.026 \\
Transfer  & TI-FGSM~\cite{dong2019evading}       & .088\,[.069,.112] & .080\,[.062,.103] & .924\,$\pm$\,.028 & .076\,[.058,.098] & .069\,[.052,.090] & .924\,$\pm$\,.028 \\
Transfer  & SINI-FGSM~\cite{Lin2020Nesterov}     & .034\,[.023,.050] & .024\,[.015,.038] & .753\,$\pm$\,.031 & .021\,[.013,.034] & .015\,[.008,.026] & .753\,$\pm$\,.031 \\
\midrule
Query     & Square~\cite{andriushchenko2020square}& .005\,[.002,.013] & .004\,[.001,.011] & .906\,$\pm$\,.071 & .003\,[.001,.010] & .003\,[.001,.009] & .906\,$\pm$\,.071 \\
Query     & SimBA-DCT~\cite{guo2019simple}       & .002\,[.000,.008] & .002\,[.000,.008] & .978\,$\pm$\,.023 & .002\,[.000,.008] & .002\,[.000,.008] & .978\,$\pm$\,.023 \\
Query     & PBO~\cite{cheng2024pbo}              & .002\,[.000,.008] & .002\,[.000,.008] & .980\,$\pm$\,.022 & .001\,[.000,.006] & .001\,[.000,.006] & .980\,$\pm$\,.022 \\
\midrule
Ensemble  & AA-PGD~\cite{croce2020reliable}      & .149\,[.124,.179] & .083\,[.064,.106] & .603\,$\pm$\,.034 & .131\,[.107,.159] & .073\,[.056,.095] & .602\,$\pm$\,.035 \\
Ensemble  & AA-Square~\cite{croce2020reliable}   & .052\,[.038,.071] & .032\,[.022,.046] & .749\,$\pm$\,.123 & .039\,[.027,.056] & .024\,[.015,.038] & .749\,$\pm$\,.123 \\
Ensemble  & MS-GAGA~\cite{msgaga2025}            & .039\,[.027,.056] & .033\,[.022,.048] & .789\,$\pm$\,.118 & .030\,[.020,.045] & .025\,[.016,.038] & .789\,$\pm$\,.118 \\
\midrule
Agentic   & RL-PPO~\cite{domico2025adversarial}  & .002\,[.000,.008] & .002\,[.000,.008] & \textbf{.981\,$\pm$\,.022} & .001\,[.000,.006] & .001\,[.000,.006] & \textbf{.981\,$\pm$\,.022} \\
\rowcolor{ARMORColor}
Agentic   & ARMOR~\cite{lee2026armor}    & .283\,[.252,.317] & .199\,[.172,.228] & .732\,$\pm$\,.158 & .259\,[.229,.292] & .181\,[.156,.210] & .729\,$\pm$\,.161 \\
\rowcolor{ARMORPPColor}
Agentic   & \textbf{ARMOR++ (ours)}              & \textbf{.321\,[.288,.355]} & \textbf{.225\,[.197,.255]} & .728\,$\pm$\,.155 & \textbf{.287\,[.256,.321]} & \textbf{.201\,[.175,.231]} & .724\,$\pm$\,.162 \\
\bottomrule
\end{tabular}
\end{adjustbox}
\end{table*}

\subsection{Evaluation Metrics and Statistical Reporting}\label{subsec:metrics}

The Attack Success Rate is $\mathrm{ASR} = \frac{1}{N} \sum_{i=1}^{N} \mathbbm{1}[\hat{y}_i \neq y_i]$. The Weighted ASR is a secondary quality-aware diagnostic, not a probability, that weights each success by perceptual similarity, $\mathrm{wASR} = \frac{1}{N} \sum_{i=1}^{N} \mathrm{SSIM}(\pmb{x}_i,\pmb{x}_{i,\mathrm{adv}})\, \mathbbm{1}[\hat{y}_i \neq y_i]$. SSIM and PSNR are reported per dataset because the clean references differ.
Each reported SSIM is the mean structural similarity between the clean image and the adversarial example selected for the corresponding detector column, averaged over the images that the column counts as successful, so columns with disjoint success subsets, or with per-detector candidate selection, need not share a single value. With $\hat{y}_i^m$ the prediction of model $m$ for image $i$, the transferability metrics are
\begin{align}
a_i
&=
\mathbbm{1}\!\left[
\bigvee_{r=1}^{R}
\hat{y}_i^{f_r}\neq y_i
\right],
\label{eq:surr_error_indicator}
\\
b_i
&=
\mathbbm{1}\!\left[
\hat{y}_i^{g}\neq y_i
\right],
\label{eq:tgt_error_indicator}
\\
p_{\mathrm{surr}}
&=
\frac{1}{N}
\sum_{i=1}^{N} a_i,
\qquad
p_{\mathrm{tgt}}
=
\frac{1}{N}
\sum_{i=1}^{N} b_i,
\label{eq:psurr_ptgt}
\\
p_{\mathrm{cond}}
&=
\frac{
\sum_{i=1}^{N} a_i b_i
}{
\sum_{i=1}^{N} a_i
}.
\label{eq:pcond}
\end{align}
The transfer gap is $\Delta\mathrm{ASR} = p_{\mathrm{surr}}-p_{\mathrm{tgt}}$. When $p_{\mathrm{surr}}\approx 1$, as for all strong surrogate-feedback methods, $p_{\mathrm{cond}}\approx p_{\mathrm{tgt}}$, so $p_{\mathrm{cond}}$ is informative chiefly for methods with incomplete surrogate success such as TI-FGSM.

With $N=713$ on AADD-LQ and $N=693$ on AADD-HQ, $95\%$ Wilson confidence intervals~\cite{wilson1927ci} are reported for ASR-type proportions and bootstrapped $95\%$ CIs ($B=10{,}000$) for wASR. Each method is run with three seeds $\{0,1,42\}$, the image-level majority vote across seeds is reported, and the Wilson CI is applied to the resulting $N$ Bernoulli trials, so the interval reflects $N$ independent images rather than $3N$ pooled trials. Between-seed variability is reported separately as $\mathrm{SD}_{\mathrm{seed}}$. Absolute risk differences with paired (McNemar-based) $95\%$ CIs are the primary effect size, and relative percentages, which are easily inflated over low baselines, are quoted only alongside the corresponding absolute difference. All pairwise tests use the exact two-sided McNemar test on per-image outcomes with Holm--Bonferroni correction over the family of $12$ baseline comparisons on AADD-LQ/ViT-B/16 (Table~\ref{tab:mcnemar}). Cells without per-image outcomes are reported with CI overlap only and are not interpreted as formally tested. For reliability interpretation, a one-sided $95\%$ Clopper--Pearson lower bound on the adversarial reliability $R(g) = 1 - \mathrm{ASR}$ is reported as $R_{\min}(g, 0.95) = 1 - U_{0.95}(N,s)$, where $U_{0.95}(N,s)$ is the Clopper--Pearson upper $95\%$ bound on the success proportion given $s$ successes in $N$ trials.

\subsection{Implementation Details and Reproducibility}\label{subsec:impl}

All adversarial optimization runs on NVIDIA RTX 4090 GPUs. The \textit{Analysis Agent} uses Qwen2.5-VL-32B-Instruct-AWQ~\cite{qwen2025vl}, and the \textit{Conductor} and \textit{Advisor} agents use Qwen3-32B-AWQ~\cite{yang2025qwen3}, hosted on AWS EC2 \texttt{g6e.8xlarge}. The default configuration is the following. CW-style: $U_{\mathrm{CW}}=1000$, $c=1.0$, $\eta_{\mathrm{CW}}=0.01$. JSMA: $B_{\mathrm{JSMA}}=2$, $U_{\mathrm{JSMA}}=500$. STA: $\theta_{\mathrm{STA}}=0.1$, $U_{\mathrm{STA}}=200$, $s_{\max}=2\,\mathrm{px}$. SSA: $Q_{\mathrm{SSA}}=20$, $U_{\mathrm{SSA}}=300$, $\rho_{\mathrm{SSA}}=0.5$, $\delta_{\min}=0.05$. BSR: $N_b=8$ ($2{\times}4$ grid), $Q_{\mathrm{BSR}}=20$, $U_{\mathrm{BSR}}=300$. Mixer: $500$ randomized hill-climbing iterations. Outer loop: $K_{\max}=5$, $W=3$, $\kappa=0.6\mu_p$, $\varrho=0.05$, $\Delta_\epsilon=2/255$, $\Delta_\tau=0.05$, $(\epsilon_{\max},\tau_{\min})=(16/255, 0.5)$. Mixing weights: $\lambda_1=1.0$, $\lambda_2=10.0$, $\lambda_3=0.1$, $\lambda_4^{(0)}=0.20$, $\rho_{\mathrm{mix}}=0.30$. Critique weights: $\mu_p=1.0$, $\mu_\ell=0.5$, $\mu_s=5.0$. Numerical floors: $\varepsilon_0=10^{-6}$, $\xi=10^{-3}$, $\epsilon_{\mathrm{tie}}=10^{-4}$. Each primitive executes its full inner loop once at $k=0$ to populate the archive, and subsequent iterations warm-start from the current master and refine at reduced step counts, so the surrogate forward-pass budget is dominated by the $k=0$ pass. A forward pass denotes one evaluation of the three-network surrogate ensemble. The decoding temperature is $0$ for the two generative LLM agents, whereas the VLM Analysis Agent uses temperature $0.2$ and top-$p=0.9$, since greedy decoding produced degenerate image-grounded JSON. All proposed hyperparameter updates are projected by Eq.~\eqref{eq:advisor_update}, so invalid settings cannot be produced.

\section{Results}\label{sec:results}

\subsection{Attack Success and Perceptual Quality}\label{subsec:main_results}

Blind-target transfer on AADD-LQ and AADD-HQ is reported in Tables~\ref{tab:lq_blind_main} and~\ref{tab:hq_blind_main}, while surrogate behavior is reported in Tables~\ref{tab:surrogate} and~\ref{tab:hq_surrogate}. All ASR values are exact proportions over the regime-specific sample size ($N=713$ for AADD-LQ and $N=693$ for AADD-HQ), and the headline result corresponds to $316/713=0.443$.

\textbf{AADD-LQ.}
The strongest blind-target transfer is obtained by ARMOR++, with ASR $0.443$ on ViT-B/16 and $0.408$ on Swin-B. Relative to ARMOR, the gains are $+4.7$ points on ViT-B/16 ($0.396\to0.443$, paired-difference $95\%$ CI $[0.019,0.077]$, exact two-sided McNemar $p\approx1.9\times10^{-3}$, significant after Holm correction) and $+3.7$ points on Swin-B ($0.371\to0.408$). Relative to AutoAttack-PGD ($0.196$ and $0.183$), the gains widen to $+24.7$ and $+22.5$ points, which confirms that the proposed orchestration improves transfer beyond stronger surrogate optimization alone. All pairwise improvements over the baselines in Table~\ref{tab:mcnemar} remain significant under the exact McNemar test after Holm correction.

\textbf{AADD-HQ.}
All methods transfer less on HQ, confirming it is the harder regime. ARMOR++ remains the strongest method, with ASR $0.321$ on ViT-B/16 and $0.287$ on Swin-B, giving $+3.8$ and $+2.8$ points over ARMOR and $+17.2$ and $+15.6$ points over AutoAttack-PGD. Because paired McNemar testing is not run on the HQ or Swin-B cells, and because the HQ margins over ARMOR are small relative to the Wilson intervals in Table~\ref{tab:hq_blind_main}, the HQ result is interpreted as a consistent ordering rather than a formally significant gain (Section~\ref{subsec:limitations}).

\textbf{Perceptual quality.}
The wASR values confirm that stronger transfer is not obtained merely by degrading image quality. Very high SSIM is preserved by query-based attacks and RL-PPO, but transfer remains rare, whereas AutoAttack-PGD transfers more often at substantially lower SSIM ($\sim 0.56$ on LQ, $\sim 0.60$ on HQ). The highest wASR on both targets and regimes is attained by ARMOR++, indicating the strongest success--fidelity balance, with $\mathrm{SD}_{\mathrm{seed}} \le 0.012$ for its ASR (full per-seed table in the supplementary material).

\begin{table}[t]
\centering
\caption{Exact two-sided McNemar tests on AADD-LQ with ViT-B/16. Discordant counts $b/c$ compare ARMOR++ with each baseline over the same $N=713$ images. All listed comparisons remain significant after Holm--Bonferroni correction.}
\label{tab:mcnemar}
\begin{adjustbox}{max width=\linewidth}
\begin{tabular}{lcc}
\toprule
\textbf{Baseline} & \textbf{$b/c$ (discordant pairs)} & \textbf{$p$-value (two-sided)} \\
\midrule
MI-FGSM       & $271/3$   & $<10^{-10}$ \\
DI-FGSM       & $290/1$   & $<10^{-10}$ \\
TI-FGSM       & $222/14$  & $<10^{-10}$ \\
SINI-FGSM     & $278/4$   & $<10^{-10}$ \\
Square        & $312/2$   & $<10^{-10}$ \\
SimBA-DCT     & $314/1$   & $<10^{-10}$ \\
PBO           & $314/0$   & $<10^{-10}$ \\
AA-PGD        & $199/23$  & $<10^{-10}$ \\
AA-Square     & $268/6$   & $<10^{-10}$ \\
MS-GAGA       & $277/4$   & $<10^{-10}$ \\
RL-PPO        & $314/0$   & $<10^{-10}$ \\
\rowcolor{ARMORColor}
ARMOR & $74/40$   & $1.9{\times}10^{-3}$ \\
\bottomrule
\end{tabular}
\end{adjustbox}
\end{table}

\begin{figure*}[!t]
    \centering
    \includegraphics[width=\linewidth]{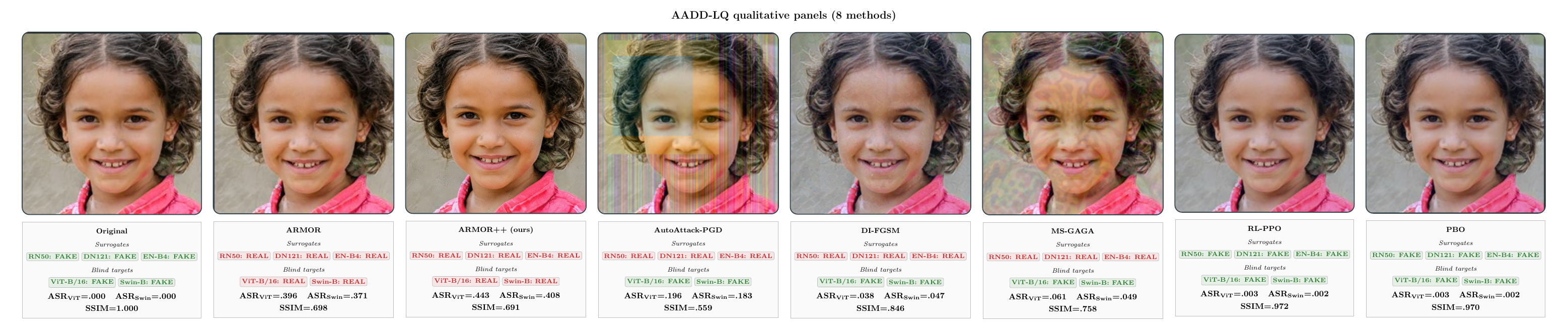}
    \caption{Representative AADD-LQ comparison across eight methods. Each panel shows one adversarial example above its per-detector predictions, with green for a correct \emph{fake} decision and red for a fooled \emph{real} decision, separated into the three CNN surrogates (ResNet-50, DenseNet-121, EfficientNet-B4) and the two blind transformer targets (ViT-B/16, Swin-B), and the footer lists the dataset-level blind-target ASR and the mean SSIM from Table~\ref{tab:lq_blind_main}.}
    \label{fig:qualitative_lq}
\end{figure*}

\begin{figure*}[!t]
    \centering
    \includegraphics[width=\linewidth]{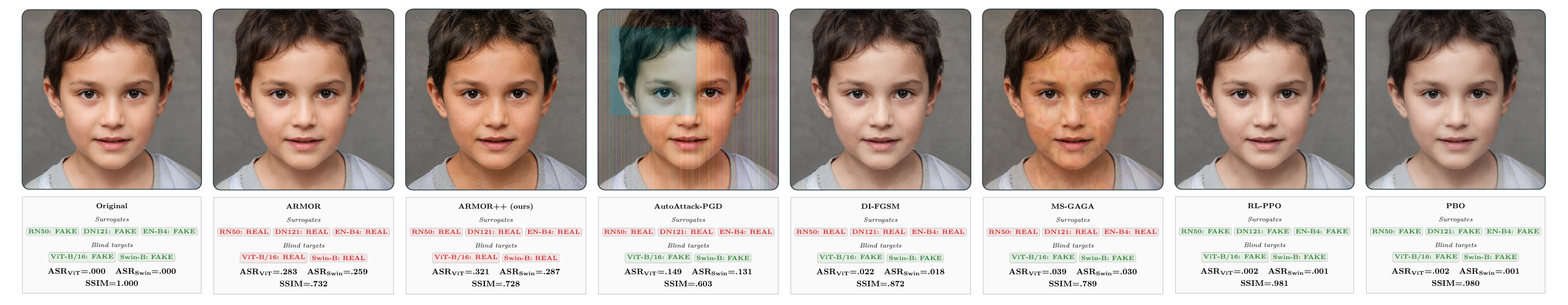}
    \caption{Representative AADD-HQ comparison using the layout of Fig.~\ref{fig:qualitative_lq}. Each panel shows one adversarial example above its per-detector predictions, with green for a correct \emph{fake} decision and red for a fooled \emph{real} decision, separated into the three CNN surrogates (ResNet-50, DenseNet-121, EfficientNet-B4) and the two blind transformer targets (ViT-B/16, Swin-B), and the footer lists the dataset-level blind-target ASR and the mean SSIM from Table~\ref{tab:hq_blind_main}.}
    \label{fig:qualitative_hq}
\end{figure*}

\textbf{Qualitative comparison.}
Figs.~\ref{fig:qualitative_lq} and~\ref{fig:qualitative_hq} present eight-method comparisons on representative AADD-LQ and AADD-HQ samples. Each panel pairs one adversarial image with its per-detector predictions, separated into the three CNN surrogates and the two blind transformer targets, where green marks a correct \emph{fake} decision and red a fooled \emph{real} decision, and the footer reports the dataset-level blind-target ASR together with the mean SSIM of Tables~\ref{tab:lq_blind_main} and~\ref{tab:hq_blind_main}. The clean image is read as \emph{fake} by all five detectors, fixing the unperturbed reference. ARMOR and ARMOR++ are the only methods that turn the entire blind-target row red, flipping both ViT-B/16 and Swin-B, and ARMOR++ achieves this while its multi-domain perturbation stays close to imperceptible, consistent with its high SSIM ($0.691$ on LQ and $0.728$ on HQ). The three strong non-agentic methods, AutoAttack-PGD, DI-FGSM, and MS-GAGA, saturate the surrogate row but leave both blind targets green, which is a direct visualization of surrogate overfitting without transfer. The contrast is sharpest for AutoAttack-PGD and MS-GAGA, whose block, stripe, and blotch artifacts are the most conspicuous distortions in the figure yet still fail to cross the architecture gap, showing that perceptual cost and blind-target transfer are decoupled and that raw surrogate distortion is not sufficient for evasion. RL-PPO and PBO occupy the opposite extreme, preserving near-original fidelity ($\mathrm{SSIM}\ge 0.97$) but flipping no detector at all. The HQ panel reproduces the same ordering with visibly subtler perturbations, in line with the smaller HQ margins of Table~\ref{tab:hq_blind_main}.

\subsection{Transferability Analysis}\label{subsec:transferability}

The metrics of Eqs.~\eqref{eq:psurr_ptgt}--\eqref{eq:pcond} are reported in Tables~\ref{tab:transfer} and~\ref{tab:transfer_hq} for AADD-LQ and AADD-HQ with ViT-B/16 as the blind target. Near-saturated surrogate success and the smallest transfer gap are obtained by ARMOR++ in both regimes, and the gap widens uniformly on HQ, consistent with the harder spectral cues of high-quality imagery.

\begin{table}[t]
\centering
\caption{Surrogate-to-target transferability on AADD-LQ with ViT-B/16. $\Delta\mathrm{ASR}=p_{\mathrm{surr}}-p_{\mathrm{tgt}}$, where lower is preferable when target ASR is high. RL-PPO has indeterminate $p_{\mathrm{cond}}$ because $p_{\mathrm{surr}}\approx 0$.}
\label{tab:transfer}
\begin{adjustbox}{max width=\linewidth}
\begin{tabular}{lcccc}
\toprule
\textbf{Method} & $\bm{p_{\mathrm{surr}}}$ $\uparrow$ & $\bm{p_{\mathrm{tgt}}}$ $\uparrow$ & $\bm{p_{\mathrm{cond}}}$ $\uparrow$ & $\bm{\Delta\mathrm{ASR}}$ $\downarrow$ \\
\midrule
TI-FGSM~\cite{dong2019evading}       & 0.914 & 0.151 & 0.165 & 0.763 \\
MI-FGSM~\cite{dong2018boosting}      & 1.000 & 0.068 & 0.068 & 0.932 \\
DI-FGSM~\cite{xie2019improving}      & 0.999 & 0.038 & 0.038 & 0.961 \\
SINI-FGSM~\cite{Lin2020Nesterov}     & 1.000 & 0.059 & 0.059 & 0.941 \\
\midrule
AA-PGD~\cite{croce2020reliable}      & 1.000 & 0.196 & 0.196 & 0.804 \\
AA-Square~\cite{croce2020reliable}   & 0.999 & 0.076 & 0.076 & 0.923 \\
MS-GAGA~\cite{msgaga2025}            & 1.000 & 0.061 & 0.061 & 0.939 \\
\midrule
RL-PPO~\cite{domico2025adversarial}  & 0.000 & 0.003 & ---   & ---   \\
\rowcolor{ARMORColor}
ARMOR~\cite{lee2026armor}    & 1.000 & 0.396 & 0.396 & 0.604 \\
\rowcolor{ARMORPPColor}
\textbf{ARMOR++ (ours)} & \textbf{1.000} & \textbf{0.443} & \textbf{0.443} & \textbf{0.557} \\
\bottomrule
\end{tabular}
\end{adjustbox}
\end{table}

\begin{table}[t]
\centering
\caption{Surrogate-to-target transferability on AADD-HQ with ViT-B/16. Gaps are larger than on AADD-LQ, but ARMOR++ keeps the smallest transfer gap.}
\label{tab:transfer_hq}
\begin{adjustbox}{max width=\linewidth}
\begin{tabular}{lcccc}
\toprule
\textbf{Method} & $\bm{p_{\mathrm{surr}}}$ $\uparrow$ & $\bm{p_{\mathrm{tgt}}}$ $\uparrow$ & $\bm{p_{\mathrm{cond}}}$ $\uparrow$ & $\bm{\Delta\mathrm{ASR}}$ $\downarrow$ \\
\midrule
TI-FGSM~\cite{dong2019evading}       & 0.873 & 0.088 & 0.101 & 0.785 \\
MI-FGSM~\cite{dong2018boosting}      & 0.997 & 0.041 & 0.041 & 0.956 \\
DI-FGSM~\cite{xie2019improving}      & 0.994 & 0.022 & 0.022 & 0.972 \\
SINI-FGSM~\cite{Lin2020Nesterov}     & 0.999 & 0.034 & 0.034 & 0.965 \\
\midrule
AA-PGD~\cite{croce2020reliable}      & 1.000 & 0.149 & 0.149 & 0.851 \\
AA-Square~\cite{croce2020reliable}   & 0.997 & 0.052 & 0.052 & 0.945 \\
MS-GAGA~\cite{msgaga2025}            & 0.999 & 0.039 & 0.039 & 0.960 \\
\midrule
RL-PPO~\cite{domico2025adversarial}  & 0.000 & 0.002 & ---   & ---   \\
\rowcolor{ARMORColor}
ARMOR~\cite{lee2026armor}    & 1.000 & 0.283 & 0.283 & 0.717 \\
\rowcolor{ARMORPPColor}
\textbf{ARMOR++ (ours)} & \textbf{1.000} & \textbf{0.321} & \textbf{0.321} & \textbf{0.679} \\
\bottomrule
\end{tabular}
\end{adjustbox}
\end{table}

\subsection{Compute Accounting}\label{subsec:compute_parity}

The per-image surrogate forward-pass count is reported in Table~\ref{tab:compute}. Compute parity with the gradient baselines is not claimed for ARMOR++, and its surrogate budget is dominated by the five inner primitive loops.
Each Phase~4 mixing step additionally evaluates $\bar p_{\mathrm{surr}}$ on candidate mixtures, and these mixer evaluations are surrogate-ensemble forward passes that are not folded into the primitive-loop count tabulated above. Separately, ARMOR++ incurs the inference cost of one $32$B VLM and one $32$B LLM per image (Section~\ref{subsec:compute}). These agents perform no surrogate forward passes, since they reason over the diagnostic vectors $\pmb d_m^{(k)}$ rather than raw images, but their wall-clock and energy cost dominates in practice, so the forward-pass count is a lower bound on total cost.

\begin{table}[t]
\centering
\caption{Per-image surrogate forward-pass count. One pass denotes one evaluation of the three-network surrogate ensemble. ARMOR++ averages $\sim13{,}400$ passes because several primitives terminate before the nominal $13{,}700$-evaluation budget.}
\label{tab:compute}
\begin{adjustbox}{max width=\linewidth}
\begin{tabular}{lr}
\toprule
\textbf{Method} & \textbf{Surrogate fwd-passes / image} \\
\midrule
MI-FGSM~\cite{dong2018boosting}      & $300$    \\
DI-FGSM~\cite{xie2019improving}      & $300$    \\
TI-FGSM~\cite{dong2019evading}       & $300$    \\
SINI-FGSM~\cite{Lin2020Nesterov}     & $300$    \\
Square~\cite{andriushchenko2020square}& $2{,}500$ \\
SimBA-DCT~\cite{guo2019simple}       & $2{,}500$ \\
PBO~\cite{cheng2024pbo}              & $2{,}500$ \\
AA-PGD~\cite{croce2020reliable}      & $\sim 5{,}000$ \\
AA-Square~\cite{croce2020reliable}   & $\sim 5{,}000$ \\
MS-GAGA~\cite{msgaga2025}            & $\sim 4{,}000$ \\
RL-PPO~\cite{domico2025adversarial}  & $\sim 1{,}500$ \\
\rowcolor{ARMORColor}
ARMOR~\cite{lee2026armor}    & $\sim 9{,}500$  \\
\rowcolor{ARMORPPColor}
\textbf{ARMOR++ (ours)}              & $\sim 13{,}400$ \\
\bottomrule
\end{tabular}
\end{adjustbox}
\end{table}

\subsection{Surrogate-Detector Results}\label{subsec:surrogate}

Surrogate performance is reported in Tables~\ref{tab:surrogate} and~\ref{tab:hq_surrogate}. As expected for surrogate-feedback attacks, strong methods reach near-perfect surrogate ASR. The discriminating quantity is therefore perceptual cost: AutoAttack-PGD reaches $\mathrm{ASR}=1.00$ at surrogate SSIM $\sim 0.56$, whereas SSIM is kept near $0.98$ by ARMOR++ at the same ASR. The EfficientNet-B4 column for ARMOR is zero-shot because that backbone is excluded from its ensemble, and the resulting gap supports the role of surrogate diversity.

\begin{table*}[t]
\centering
\caption{Surrogate-detector performance on AADD-LQ. ASR is shown with $95\%$ Wilson CIs, and SSIM is mean $\pm$ one standard deviation. EfficientNet-B4 is zero-shot for ARMOR.}
\label{tab:surrogate}
\begin{adjustbox}{max width=\linewidth}
\begin{tabular}{llcccccc}
\toprule
\multirow{2}{*}{\textbf{Category}} & \multirow{2}{*}{\textbf{Method}} &
\multicolumn{2}{c}{\textbf{ResNet-50}} & \multicolumn{2}{c}{\textbf{DenseNet-121}} & \multicolumn{2}{c}{\textbf{EfficientNet-B4}} \\
\cmidrule(lr){3-4}\cmidrule(lr){5-6}\cmidrule(lr){7-8}
& & \textbf{ASR} & \textbf{SSIM} & \textbf{ASR} & \textbf{SSIM} & \textbf{ASR} & \textbf{SSIM} \\
\midrule
Transfer  & MI-FGSM~\cite{dong2018boosting}        & .994\,[.985,.998] & .743\,$\pm$\,.038 & 1.000\,[.995,1.000] & .743\,$\pm$\,.038 & .989\,[.978,.995] & .743\,$\pm$\,.038 \\
Transfer  & DI-FGSM~\cite{xie2019improving}        & .906\,[.882,.926] & .846\,$\pm$\,.033 & .997\,[.990,.999]   & .846\,$\pm$\,.033 & .902\,[.877,.922] & .846\,$\pm$\,.033 \\
Transfer  & TI-FGSM~\cite{dong2019evading}         & .713\,[.679,.745] & .904\,$\pm$\,.036 & .903\,[.879,.923]   & .904\,$\pm$\,.036 & .701\,[.667,.733] & .904\,$\pm$\,.036 \\
Transfer  & SINI-FGSM~\cite{Lin2020Nesterov}       & .935\,[.915,.951] & .711\,$\pm$\,.039 & 1.000\,[.995,1.000] & .711\,$\pm$\,.039 & .918\,[.896,.936] & .711\,$\pm$\,.039 \\
\midrule
Query     & Square~\cite{andriushchenko2020square} & .000\,[.000,.005] & .881\,$\pm$\,.084 & .023\,[.014,.038]   & .881\,$\pm$\,.084 & .000\,[.000,.005] & .881\,$\pm$\,.084 \\
Query     & SimBA-DCT~\cite{guo2019simple}         & .154\,[.129,.183] & .968\,$\pm$\,.031 & .176\,[.149,.207]   & .968\,$\pm$\,.031 & .142\,[.118,.171] & .968\,$\pm$\,.031 \\
Query     & PBO~\cite{cheng2024pbo}                & .000\,[.000,.005] & .970\,$\pm$\,.030 & .000\,[.000,.005]   & .970\,$\pm$\,.030 & .000\,[.000,.005] & .970\,$\pm$\,.030 \\
\midrule
Ensemble  & AA-PGD~\cite{croce2020reliable}        & 1.000\,[.995,1.000] & .551\,$\pm$\,.036 & 1.000\,[.995,1.000] & .571\,$\pm$\,.038 & 1.000\,[.995,1.000] & .560\,$\pm$\,.037 \\
Ensemble  & AA-Square~\cite{croce2020reliable}     & .976\,[.962,.985] & .725\,$\pm$\,.142 & .959\,[.942,.972] & .716\,$\pm$\,.137 & .968\,[.953,.979] & .722\,$\pm$\,.139 \\
Ensemble  & MS-GAGA~\cite{msgaga2025}              & 1.000\,[.995,1.000] & .758\,$\pm$\,.131 & 1.000\,[.995,1.000] & .758\,$\pm$\,.131 & .997\,[.990,.999] & .758\,$\pm$\,.131 \\
\midrule
Agentic   & RL-PPO~\cite{domico2025adversarial}    & .000\,[.000,.005] & .972\,$\pm$\,.030 & .000\,[.000,.005] & .972\,$\pm$\,.030 & .000\,[.000,.005] & .972\,$\pm$\,.030 \\
\rowcolor{ARMORColor}
Agentic   & ARMOR~\cite{lee2026armor}      & 1.000\,[.995,1.000] & .968\,$\pm$\,.042 & 1.000\,[.995,1.000] & .961\,$\pm$\,.047 & .742\,[.708,.773] & .968\,$\pm$\,.044 \\
\rowcolor{ARMORPPColor}
Agentic   & \textbf{ARMOR++ (ours)}                & \textbf{1.000\,[.995,1.000]} & \textbf{.981\,$\pm$\,.033} & \textbf{1.000\,[.995,1.000]} & \textbf{.976\,$\pm$\,.040} & \textbf{1.000\,[.995,1.000]} & \textbf{.978\,$\pm$\,.037} \\
\bottomrule
\end{tabular}
\end{adjustbox}
\end{table*}

\begin{table*}[t]
\centering
\caption{Surrogate-detector performance on AADD-HQ. ARMOR++ again reaches near-perfect surrogate ASR with the highest SSIM among transferring methods.}
\label{tab:hq_surrogate}
\begin{adjustbox}{max width=\linewidth}
\begin{tabular}{llcccccc}
\toprule
\multirow{2}{*}{\textbf{Category}} & \multirow{2}{*}{\textbf{Method}} &
\multicolumn{2}{c}{\textbf{ResNet-50}} & \multicolumn{2}{c}{\textbf{DenseNet-121}} & \multicolumn{2}{c}{\textbf{EfficientNet-B4}} \\
\cmidrule(lr){3-4}\cmidrule(lr){5-6}\cmidrule(lr){7-8}
& & \textbf{ASR} & \textbf{SSIM} & \textbf{ASR} & \textbf{SSIM} & \textbf{ASR} & \textbf{SSIM} \\
\midrule
Transfer  & MI-FGSM~\cite{dong2018boosting}        & .981\,[.967,.989] & .784\,$\pm$\,.029 & .997\,[.990,.999] & .784\,$\pm$\,.029 & .974\,[.959,.984] & .784\,$\pm$\,.029 \\
Transfer  & DI-FGSM~\cite{xie2019improving}        & .872\,[.846,.895] & .872\,$\pm$\,.026 & .991\,[.982,.996] & .872\,$\pm$\,.026 & .868\,[.842,.891] & .872\,$\pm$\,.026 \\
Transfer  & TI-FGSM~\cite{dong2019evading}         & .668\,[.633,.701] & .924\,$\pm$\,.028 & .879\,[.854,.901] & .924\,$\pm$\,.028 & .652\,[.617,.686] & .924\,$\pm$\,.028 \\
Transfer  & SINI-FGSM~\cite{Lin2020Nesterov}       & .921\,[.900,.939] & .753\,$\pm$\,.031 & .997\,[.990,.999] & .753\,$\pm$\,.031 & .899\,[.876,.918] & .753\,$\pm$\,.031 \\
\midrule
Query     & Square~\cite{andriushchenko2020square} & .000\,[.000,.005] & .906\,$\pm$\,.071 & .017\,[.009,.030] & .906\,$\pm$\,.071 & .000\,[.000,.005] & .906\,$\pm$\,.071 \\
Query     & SimBA-DCT~\cite{guo2019simple}         & .117\,[.094,.143] & .978\,$\pm$\,.023 & .142\,[.118,.171] & .978\,$\pm$\,.023 & .107\,[.085,.133] & .978\,$\pm$\,.023 \\
Query     & PBO~\cite{cheng2024pbo}                & .000\,[.000,.005] & .980\,$\pm$\,.022 & .000\,[.000,.005] & .980\,$\pm$\,.022 & .000\,[.000,.005] & .980\,$\pm$\,.022 \\
\midrule
Ensemble  & AA-PGD~\cite{croce2020reliable}        & 1.000\,[.995,1.000] & .595\,$\pm$\,.033 & 1.000\,[.995,1.000] & .613\,$\pm$\,.035 & 1.000\,[.995,1.000] & .601\,$\pm$\,.034 \\
Ensemble  & AA-Square~\cite{croce2020reliable}     & .961\,[.944,.973] & .755\,$\pm$\,.121 & .942\,[.922,.957] & .748\,$\pm$\,.118 & .953\,[.935,.967] & .751\,$\pm$\,.120 \\
Ensemble  & MS-GAGA~\cite{msgaga2025}              & .999\,[.993,1.000] & .789\,$\pm$\,.118 & .999\,[.993,1.000] & .789\,$\pm$\,.118 & .993\,[.984,.997] & .789\,$\pm$\,.118 \\
\midrule
Agentic   & RL-PPO~\cite{domico2025adversarial}    & .000\,[.000,.005] & .981\,$\pm$\,.022 & .000\,[.000,.005] & .981\,$\pm$\,.022 & .000\,[.000,.005] & .981\,$\pm$\,.022 \\
\rowcolor{ARMORColor}
Agentic   & ARMOR~\cite{lee2026armor}      & 1.000\,[.995,1.000] & .974\,$\pm$\,.036 & 1.000\,[.995,1.000] & .968\,$\pm$\,.041 & .703\,[.668,.736] & .974\,$\pm$\,.038 \\
\rowcolor{ARMORPPColor}
Agentic   & \textbf{ARMOR++ (ours)}                & \textbf{1.000\,[.995,1.000]} & \textbf{.985\,$\pm$\,.029} & \textbf{1.000\,[.995,1.000]} & \textbf{.981\,$\pm$\,.034} & \textbf{1.000\,[.995,1.000]} & \textbf{.983\,$\pm$\,.031} \\
\bottomrule
\end{tabular}
\end{adjustbox}
\end{table*}

\subsection{Ablation Study}\label{subsec:ablation}

A component-removal analysis is provided in Table~\ref{tab:ablation} on AADD-LQ with ViT-B/16. The first two rows fix reference points, and the lettered rows quantify the sensitivity of the result to individual components.

\begin{table}[t]
\centering
\caption{Ablation on AADD-LQ with ViT-B/16. Rows compare component removals under the same no-query protocol. Row (h) isolates the third surrogate, and ASR is reported with $95\%$ Wilson CIs.}
\label{tab:ablation}
\begin{adjustbox}{max width=\linewidth}
\begin{tabular}{lccc}
\toprule
\textbf{Configuration} & \textbf{ASR [95\% CI]} & \textbf{wASR} & \textbf{SSIM} \\
\midrule
\rowcolor{ARMORPPColor}
\textbf{ARMOR++ (Full)} & \textbf{.443\,[.407,.480]} & \textbf{.312} & .691\,$\pm$\,.169 \\
\rowcolor{ARMORColor}
ARMOR~\cite{lee2026armor} (baseline) & .396\,[.361,.432] & .280 & .698\,$\pm$\,.173 \\
\midrule
(a) Three primitives, 3 surrogates, full agents & .396\,[.361,.432] & .280 & .698\,$\pm$\,.173 \\
(b) Five primitives, uniform mixing (no Mixer)   & .104\,[.084,.129] & .095 & .919\,$\pm$\,.041 \\
(c) Five primitives, no Advisor                  & .287\,[.256,.321] & .202 & .706\,$\pm$\,.152 \\
(d) No Analysis Agent (uniform mask)             & .218\,[.190,.250] & .152 & .695\,$\pm$\,.162 \\
(e) No Conductor (fixed defaults)                & .311\,[.279,.346] & .218 & .700\,$\pm$\,.161 \\
(f) No Strategist (no stagnation escalation)     & .374\,[.340,.410] & .263 & .703\,$\pm$\,.171 \\
(g) No agentic components                        & .012\,[.006,.024] & .011 & .945\,$\pm$\,.031 \\
\midrule
(h) Two surrogates (RN50+DN121 only)             & .421\,[.385,.457] & .297 & .692\,$\pm$\,.171 \\
(i) No entropy regularizer ($\lambda_4=0$)       & .402\,[.367,.438] & .283 & .706\,$\pm$\,.161 \\
(j) SSA removed                                  & .388\,[.353,.424] & .272 & .697\,$\pm$\,.171 \\
(k) BSR removed                                  & .378\,[.344,.414] & .265 & .700\,$\pm$\,.172 \\
\bottomrule
\end{tabular}
\end{adjustbox}
\end{table}

The most severe degradation is observed when adaptive mixing is replaced by uniform averaging (row (b), ASR $0.104$). Removing the Analysis Agent (row (d)) reduces ASR to $0.218$, the Advisor (row (c)) to $0.287$, and the Strategist (row (f)) to $0.374$. Row (a) reaches $0.396$, so the gain from $0.396$ to $0.443$ is associated with adding SSA and BSR, the third surrogate, and the entropy regularizer. Their individual removals cost $12.4\%$ (row (j)), $14.7\%$ (row (k)), $5.0\%$ (row (h)), and $9.3\%$ (row (i)) in relative ASR, identifying the proposed multi-domain design as the main source of the gain. Two cautions apply. Rows (b) and (g) use fixed default hyperparameters, and row (g) ($0.012$) falls below even single-method transfer (MI-FGSM $=0.068$), so the table should be read as a component-removal study rather than as proof of dominance over every statically tuned ensemble. Several lettered rows also differ by less than their Wilson half-widths, so the attributions in rows (h)--(k) are ordinal rather than independently significant.

\subsection{Cross-Dataset Robustness}\label{subsec:crossdataset}

ARMOR++ and ARMOR were run unchanged on a $200$-image fake subset of DFDC-Preview~\cite{dolhansky2019dfdc}, a strict zero-shot evaluation across data distribution (Table~\ref{tab:crossdataset}). The method ordering is preserved (ARMOR++ $>$ ARMOR $>$ AA-PGD on both targets). At $N=200$ the ARMOR++-over-ARMOR margin is small ($+3.5$ points on ViT-B/16, $0.270\to0.305$) and its Wilson interval $[0.244,0.371]$ overlaps that of ARMOR $[0.213,0.336]$, so this difference is not significant at this sample size and no significant cross-dataset improvement over ARMOR is claimed. The experiment therefore supports only that a large, CI-separable margin over the non-agentic baseline is retained off-benchmark by ARMOR++.

\begin{table}[t]
\centering
\caption{Zero-shot transfer on DFDC-Preview ($N=200$). AADD-2025 surrogates and targets are used without fine-tuning or retuning, and ASR is shown with Wilson $95\%$ CIs.}
\label{tab:crossdataset}
\begin{adjustbox}{max width=\linewidth}
\begin{tabular}{lcc}
\toprule
\textbf{Method} & \textbf{ASR on ViT-B/16} & \textbf{ASR on Swin-B} \\
\midrule
MI-FGSM~\cite{dong2018boosting}      & .045\,[.022,.087] & .035\,[.016,.075] \\
AA-PGD~\cite{croce2020reliable}      & .135\,[.094,.190] & .115\,[.078,.167] \\
MS-GAGA~\cite{msgaga2025}            & .050\,[.026,.094] & .040\,[.019,.081] \\
\rowcolor{ARMORColor}
ARMOR~\cite{lee2026armor}    & .270\,[.213,.336] & .250\,[.195,.314] \\
\rowcolor{ARMORPPColor}
\textbf{ARMOR++ (ours)}              & \textbf{.305\,[.244,.371]} & \textbf{.275\,[.217,.341]} \\
\bottomrule
\end{tabular}
\end{adjustbox}
\end{table}

\subsection{Defense-Aware Reliability Evaluation}\label{subsec:defense}

ARMOR++ and ARMOR are evaluated against two representative non-adaptive defenses. The defenses are not tuned with knowledge of ARMOR++, and ARMOR++ is not re-optimized against them, so a fully adaptive evaluation (for example, expectation-over-transformation against the randomized defense) is required before any broader robustness claim and is outside the present scope (Section~\ref{subsec:limitations}). D1 is adversarial training~\cite{madry2018towards}, in which the ViT-B/16 target is retrained with PGD-$\ell_\infty$ examples ($\epsilon=8/255$, $10$ inner steps, $\alpha=1/255$, $50\%$ adversarial per batch, regenerated each epoch, $30$ epochs, final clean accuracy $94.6\%$). D2 is randomized input transformation~\cite{xie2017mitigating}, in which the input is randomly resized to $[200,224]$ and zero-padded back to $224\times 224$ with predictions averaged over $5$ draws.

Blind-target ASR under both defenses is reported with $95\%$ Wilson CIs in Tables~\ref{tab:defense} and~\ref{tab:defense_hq}. Under PGD adversarial training on AADD-LQ, $\mathrm{ASR}=0.198\,[.171,.229]$ is retained by ARMOR++, whereas the strongest non-agentic baseline (AutoAttack-PGD) drops to $0.061$. Under randomized resize, the corresponding values are $0.276$ and $0.094$, showing that the proposed attack remains the most effective stress test after non-adaptive defenses.

\begin{table}[t]
\centering
\caption{Blind-target ASR under defenses on AADD-LQ with ViT-B/16. D1 is PGD adversarial training, and D2 is randomized input transformation. ARMOR++ retains the largest residual ASR.}
\label{tab:defense}
\begin{adjustbox}{max width=\linewidth}
\begin{tabular}{lccc}
\toprule
\textbf{Method} & \textbf{Undefended} & \textbf{D1 (AdvTrain)} & \textbf{D2 (Rand.\ Resize)} \\
\midrule
MI-FGSM~\cite{dong2018boosting}      & .068\,[.052,.089] & .021\,[.013,.034] & .034\,[.023,.050] \\
AA-PGD~\cite{croce2020reliable}      & .196\,[.168,.228] & .061\,[.046,.082] & .094\,[.075,.117] \\
MS-GAGA~\cite{msgaga2025}            & .061\,[.046,.082] & .018\,[.010,.031] & .029\,[.019,.044] \\
\rowcolor{ARMORColor}
ARMOR~\cite{lee2026armor}    & .396\,[.361,.432] & .174\,[.149,.205] & .241\,[.212,.273] \\
\rowcolor{ARMORPPColor}
\textbf{ARMOR++ (ours)}              & \textbf{.443\,[.407,.480]} & \textbf{.198\,[.171,.229]} & \textbf{.276\,[.245,.310]} \\
\bottomrule
\end{tabular}
\end{adjustbox}
\end{table}

\begin{table}[t]
\centering
\caption{Blind-target ASR under defenses on AADD-HQ with ViT-B/16. The AADD-LQ ordering is preserved, and ARMOR++ retains the largest residual ASR after defense.}
\label{tab:defense_hq}
\begin{adjustbox}{max width=\linewidth}
\begin{tabular}{lccc}
\toprule
\textbf{Method} & \textbf{Undefended} & \textbf{D1 (AdvTrain)} & \textbf{D2 (Rand.\ Resize)} \\
\midrule
MI-FGSM~\cite{dong2018boosting}      & .041\,[.029,.058] & .013\,[.007,.024] & .021\,[.013,.034] \\
AA-PGD~\cite{croce2020reliable}      & .149\,[.124,.179] & .045\,[.032,.063] & .069\,[.053,.090] \\
MS-GAGA~\cite{msgaga2025}            & .039\,[.027,.056] & .011\,[.005,.022] & .018\,[.010,.031] \\
\rowcolor{ARMORColor}
ARMOR~\cite{lee2026armor}    & .283\,[.252,.317] & .124\,[.101,.151] & .172\,[.146,.202] \\
\rowcolor{ARMORPPColor}
\textbf{ARMOR++ (ours)}              & \textbf{.321\,[.288,.355]} & \textbf{.142\,[.119,.170]} & \textbf{.198\,[.171,.229]} \\
\bottomrule
\end{tabular}
\end{adjustbox}
\end{table}

\subsection{Compute Cost}\label{subsec:compute}

A single ARMOR++ run per image requires about $75$ seconds on one RTX~4090 for the attack stage, excluding VLM/LLM latency. For $713$ (AADD-LQ) $+$ $693$ (AADD-HQ) $=1{,}406$ images $\times$ $3$ seeds $=4{,}218$ runs, approximately $\sim 88$ GPU-hours of attack compute plus $\sim 30$ GPU-hours of VLM/LLM inference were consumed by ARMOR++, and the full study, including baselines, ablations, and defenses, consumed $\sim 540$ GPU-hours. Per image, the pipeline issues $2+K_{\max}=7$ generative calls: one VLM Analysis call, one Conductor call, and $K_{\max}=5$ Advisor calls, each returning hyperparameter increments for all five primitives. The Strategist and Mixer are deterministic and issue no generative calls.

\section{Discussion}\label{sec:discussion}

\subsection{Why ARMOR++ Transfers Better}

The results across AADD-LQ and AADD-HQ support the conclusion that transferable attacks against deepfake detectors benefit from coverage across multiple perturbation domains. Conventional transfer attacks optimize spatial gradients on a single surrogate family, and the resulting perturbations overfit to surrogate-specific local features. A five-primitive space is combined in ARMOR++ with three architecturally diverse surrogates, and SSA and BSR supply the complementary frequency-domain and block-structured modes whose removal costs $12$--$15\%$ relative ASR in the ablation. The comparison between the three-primitive configuration and full ARMOR++ ($0.396\to0.443$) supports the empirical value of enlarging the perturbation set and adding the third surrogate together with entropy-regularized mixing under the evaluated matched-envelope protocol. This benefit is important in the HQ regime, where forensic cues are subtle and distributed, because no single artifact type is required by ARMOR++.

\subsection{Reliability-Engineering Interpretation}\label{subsec:reliability_engineering}

The adversarial reliability of a detector $g$ under a threat distribution $\mathcal{T}$ is
\begin{equation}
R(g,\mathcal{T}) = 1 - \E_{(\pmb{x},y)\sim\mathcal{D},\,A\sim\mathcal{T}} \bigl[\Pr(\hat{y}_g(A(\pmb{x}))\neq y)\bigr],
\label{eq:reliability_def}
\end{equation}
where $A$ is an attack draw from $\mathcal{T}$, $\mathcal{D}$ the test distribution, and $\hat{y}_g$ the predicted label (for the deterministic, seeded ARMOR++ attack, $\mathcal{T}$ is the seed mixture). Under ARMOR++ on AADD-LQ with ViT-B/16, $R(g) = 1 - 0.443 = 0.557$, with a one-sided $95\%$ Clopper--Pearson lower bound $R_{\min}(g,0.95) = 0.525$ computed for $316/713$, whereas ARMOR gives $R = 0.604$ and AutoAttack-PGD $R = 0.804$. For biometric authentication, digital-evidence verification, or automated misinformation screening, a residual failure rate of $\sim 44\%$ under a feasible no-query threat model is operationally significant. The two defenses raise reliability only to $R=0.802$ (PGD adversarial training) and $R=0.724$ (randomized resize), still short of the $R\ge 0.90$ typically required for automated decisions, which indicates that high clean accuracy and standard non-adaptive defenses do not by themselves certify detector dependability under cross-architecture adversarial transfer.

\subsection{Failure-Mode Analysis}\label{subsec:failure_mode}

For the $713-316=397$ AADD-LQ images on which ViT-B/16 is not flipped by ARMOR++, each failure is associated with the primitive carrying the largest mixing weight at $K_{\max}$ (Table~\ref{tab:failure_mode}), this is an association rather than a causal attribution. Frequency-domain failures (SSA-dominated mixtures) account for $43.1\%$ of cases, indicating that high-quality images with subtle spectral artifacts are the dominant failure mode, consistent with the smaller HQ-regime gain. Block-structured failures account for $24.7\%$, predominantly very smooth synthetic faces on which block shuffling cannot find a discriminating partition, and geometric (STA) failures, at $13.6\%$, correspond to images on which the $\ell_\infty$ projection destroys the coherent warp (Section~\ref{subsubsec:sta}).

\begin{table}[t]
\centering
\caption{Failure-mode taxonomy on AADD-LQ with ViT-B/16. Each unsuccessful image is associated with the primitive carrying the largest final mixing weight, which is not a causal attribution.}
\label{tab:failure_mode}
\begin{adjustbox}{max width=\linewidth}
\begin{tabular}{lcc}
\toprule
\textbf{Dominant primitive at failure} & \textbf{Count} & \textbf{Share (\%)} \\
\midrule
SSA (frequency-domain)        & $171$ & $43.1$ \\
BSR (block-structured)        & $98$  & $24.7$ \\
CW-style (dense $\ell_2$)     & $58$  & $14.6$ \\
STA (geometric warp)          & $54$  & $13.6$ \\
JSMA (sparse saliency)        & $16$  & $4.0$  \\
\midrule
\textbf{Total failures}       & $\mathbf{397}$ & $\mathbf{100.0}$ \\
\bottomrule
\end{tabular}
\end{adjustbox}
\end{table}

\subsection{LQ versus HQ Behavior}

On AADD-LQ, artifacts are stronger and more localized, and all baselines are outperformed by ARMOR++. On AADD-HQ, the absolute ASR of every method decreases, but ARMOR++ preserves its ranking across both targets and its margin over ARMOR ($+3.8$ points on ViT-B/16, $+2.8$ on Swin-B), which indicates that the five-primitive space and closed-loop reparameterization remain useful when forensic evidence is subtle and distributed.

\subsection{Limitations and Reproducibility}\label{subsec:limitations}

The boundaries of the present evidence are stated as follows. \emph{First}, the ablation is a matched-budget component-removal analysis, not an exhaustive hyperparameter search over every static mixture of the five primitives, so the paper claims improvement over the evaluated baselines and ablations rather than universal dominance over all static controllers. \emph{Second}, exact McNemar testing is applied only to the AADD-LQ/ViT-B/16 comparisons with paired per-image outcomes, and the AADD-HQ, Swin-B, DFDC-Preview, and defense cells are reported with confidence intervals and treated as consistent orderings rather than formally established improvements. \emph{Third}, the two defenses are non-adaptive and do not establish robustness against defenses optimized with knowledge of ARMOR++ or against expectation-over-transformation attacks. \emph{Fourth}, all five backbones use ImageNet pretraining, which may introduce a shared representation channel between surrogates and targets, and a future negative control with an independently initialized target would help bound it. \emph{Fifth}, the qualitative comparisons in Figs.~\ref{fig:qualitative_lq}--\ref{fig:qualitative_hq} are illustrative and carry no statistical weight, so all quantitative claims rest on the tabulated ASR, wASR, SSIM, confidence intervals, and paired tests. For reproducibility, the camera-ready artifact release will include the detector checkpoints, the exact AADD-2025 subset indices, the per-image prediction and contingency tables behind every ASR cell, the verbatim agent prompts and decoding configurations, the attack code with fixed seeds, and the scripts that compute all intervals, tests, and reliability bounds.

\section{Conclusion}\label{sec:conclusion}

ARMOR++ was introduced as a vision-language-guided framework in which a five-primitive, multi-domain perturbation set (CW-style, JSMA, STA, SSA, BSR) is orchestrated over a three-CNN surrogate ensemble under a strict no-query protocol. Surrogate feedback alone drives optimization, mixing, reparameterization, and final selection. On AADD-LQ with ViT-B/16, ARMOR++ attains ASR $0.443$ and improves over ARMOR by $+4.7$ points ($0.396\to0.443$, paired-difference $95\%$ CI $[0.019,0.077]$, exact two-sided McNemar $p\approx1.9\times10^{-3}$ after Holm correction) and over AutoAttack-PGD by $+24.7$ points. The same ordering holds on Swin-B and in the HQ regime, with the smaller HQ and DFDC-Preview margins reported as consistent orderings rather than formal significance claims, and under two non-adaptive defenses ARMOR++ retains residual ASR of $0.198$ and $0.276$. The component-removal study indicates that semantic guidance, adaptive reparameterization, stagnation handling, entropy-regularized mixing, SSA, BSR, and surrogate diversity each affect transfer under the evaluated protocol. Taken together, the results show that current deepfake detectors retain an operationally significant residual failure rate under semantically informed, multi-domain no-query transfer, so high clean accuracy and standard non-adaptive defenses are insufficient to certify detector dependability under cross-architecture transfer. Future work should extend the evaluation to adaptive defenses, broader and independently pretrained target families, larger cross-dataset tests, and more extensive static-controller sweeps.

\bibliographystyle{IEEEtran}
\bibliography{references}


\end{document}